\documentclass{article}

\PassOptionsToPackage{numbers, compress}{natbib}

\usepackage[preprint]{neurips_2023}

\usepackage[utf8]{inputenc} 
\usepackage[T1]{fontenc}    
\usepackage{hyperref}       
\usepackage{url}            
\usepackage{booktabs}       
\usepackage{amsfonts}       
\usepackage{nicefrac}       
\usepackage{microtype}      
\usepackage{xcolor}         

\usepackage[color=red]{todonotes}
\usepackage{arydshln}
\usepackage{subcaption}
\usepackage{multirow}
\usepackage{comment}
\usepackage{colortbl}
\usepackage{tcolorbox}

\usepackage{textcomp}
\usepackage{threeparttable}
\usepackage[normalem]{ulem}
\usepackage{enumitem}
\usepackage{inconsolata}
\usepackage{CJKutf8}
\usetikzlibrary{patterns}

\usepackage{wrapfig}
\usepackage{pgfplots}
\usepgfplotslibrary{groupplots}

\usepackage[main=english,russian]{babel}

\definecolor{myYellow}{rgb}{0.9,0.9,1}

\definecolor{WindowsBlue}{RGB}{2,152,219}

\definecolor{battleshipgrey}{rgb}{0.3, 0.3, 0.3}
\definecolor{brilliantrose}{rgb}{1.0, 0.33, 0.64}
\definecolor{americanrose}{rgb}{1.0, 0.01, 0.24}
\definecolor{jweigreen}{rgb}{0,0.45,0.24}
\definecolor{bluegray}{rgb}{0.1, 0.1, 0.4}
\definecolor{ao(english)}{rgb}{0.0, 0.5, 0.0}
\definecolor{blanchedalmond}{rgb}{1.0, 0.92, 0.8}
\definecolor{atomictangerine}{rgb}{1.0, 0.6, 0.4}
\definecolor{chocolate(web)}{rgb}{0.82, 0.41, 0.12}
\definecolor{bananayellow}{rgb}{1.0, 0.88, 0.21}
\definecolor{goldenbrown}{rgb}{0.6, 0.4, 0.08}
\definecolor{aliceblue}{rgb}{0.94, 0.97, 1.0}
\definecolor{beige}{rgb}{0.96, 0.96, 0.86}
\definecolor{babyblue}{rgb}{0.54, 0.81, 0.94}
\definecolor{camel}{rgb}{0.76, 0.6, 0.42}
\definecolor{cinnamon}{rgb}{0.82, 0.41, 0.12}

\title{Deliberate then Generate: Enhanced Prompting Framework for Text Generation}

\author{%
  Bei Li$^1$\thanks{Equal Contribution. Work done when Bei Li is interning at Microsoft Research Asia.}, Rui Wang$^2$\footnotemark[1], Junliang Guo$^2$\footnotemark[1], Kaitao Song$^2$, Xu Tan$^2$\thanks{Corresponding Author}, Hany Hassan$^3$\\
  \textbf{Arul Menezes$^3$,} \textbf{Tong Xiao$^{1,4}$\footnotemark[2],} \textbf{Jiang Bian$^2$} and \textbf{JingBo Zhu$^{1,4}$} \\
  $^{1}$School of Computer Science and Engineering, Northeastern University, Shenyang, China\\
  $^{2}$Microsoft Research Asia, $^{3}$Microsoft Azure Translation,
  $^{4}$NiuTrans Research\\
  \texttt{libei\_neu@outlook.com}, \texttt{\{xiaotong,zhujingbo\}@mail.neu.edu.cn}\\
  \texttt{\{ruiwa,junliangguo,kaitaosong,xuta,hanyh,arulm,jiabia\}@microsoft.com}\\
  }
  
\begin{document}

\maketitle

\begin{abstract}


Large language models (LLMs) have shown remarkable success across a wide range of natural language generation tasks, where proper prompt designs make great impacts. While existing prompting methods are normally restricted to providing correct information, in this paper, we encourage the model to deliberate by proposing a novel Deliberate then Generate (DTG) prompting framework, which consists of error detection instructions and candidates that may contain errors. DTG is a simple yet effective technique that can be applied to various text generation tasks with minimal modifications. We conduct extensive experiments on 20+ datasets across 7 text generation tasks, including summarization, translation, dialogue, and more. We show that DTG consistently outperforms existing prompting methods and achieves state-of-the-art performance on multiple text generation tasks. We also provide in-depth analyses to reveal the underlying mechanisms of DTG, which may inspire future research on prompting for LLMs.

\end{abstract}

\section{Introduction}

Large language models~(LLMs)~\cite{brown2020language,openai2023gpt4,touvron2023llama} are revolutionizing the area of natural language generation, which have demonstrated exceptional abilities in generating coherent and fluent text as well as exhibited a remarkable aptitude in performing a diverse range of text generation tasks with high accuracy~\cite{hendy2023good,nori2023capabilities}. When adapting to downstream tasks, traditional fine-tuning methods require access to the parameters of LLMs, which hinder their application on powerful black-box LLMs~(e.g., ChatGPT) that only provide APIs to interact with. Therefore, prompting methods that guide the generation results by providing several task-specific instructions and demonstrations have attracted lots of attention in recent works~\citep{schick2020exploiting,sanh2021multitask}, which show that the prompt can significantly influence the resulting outcomes and thus require careful design.

While prompting is itself a general approach, the current use of this approach is a bit rigid, say, an LLM only operates on the basis of what is correct~\cite{brown2020language,hendy2023good,jasonwei2022CoT}. This is not the case for language acquisition where a human can learn from both positive and negative feedback and improve the ability of language use through corrections. In this work, we examine whether and how the deliberation ability emerges by asking the LLMs to rethink and learn to detect potential errors in their output. To do this, we develop a new prompting template termed Deliberate then Generate (DTG) that contains instructions and candidate outputs to enable an error detection process before generation, i.e., adding ``\textit{Please detect the error type firstly, and provide the refined results then}'' in the prompt.

A key design aspect of DTG is how to determine the candidate. One straightforward choice is utilizing the results from an extra baseline system, which typically exhibits high quality and requires only minor adjustments. Accordingly, it cannot well facilitate the deliberation ability. In this work, we propose to utilize the text that is irrelevant from the reference (e.g., such as a randomly sampled text or even an \emph{empty string}) as the candidate. In this way, the method successfully triggers the deliberation ability of LLMs, without having to resort to other text generation systems to create correction examples, which enables DTG to be easily applied to a wide range of text generation tasks only with minimal modifications in prompts.
This work is in part motivated from a psychological perspective by considering \textit{negative evidence} in developing language abilities, which is a canonical case for language learning~\citep{marcus1993negative}.

We conduct extensive experiments on 7 text generation tasks and more than 20 datasets on GPT3.5 (\texttt{text-davinci-003}) and GPT4, where the proposed DTG prompting consistently improves model performance compared to conventional prompts. GPT with DTG prompting achieves state-of-the-art performance on multiple datasets across different text generation tasks, including machine translation, simplification and commonsense generation. 
Extensive ablation studies and error statistical analysis illustrate that the proposed DTG prompting does enable deliberation ability and error avoidance before generation. 

\usetikzlibrary{backgrounds}
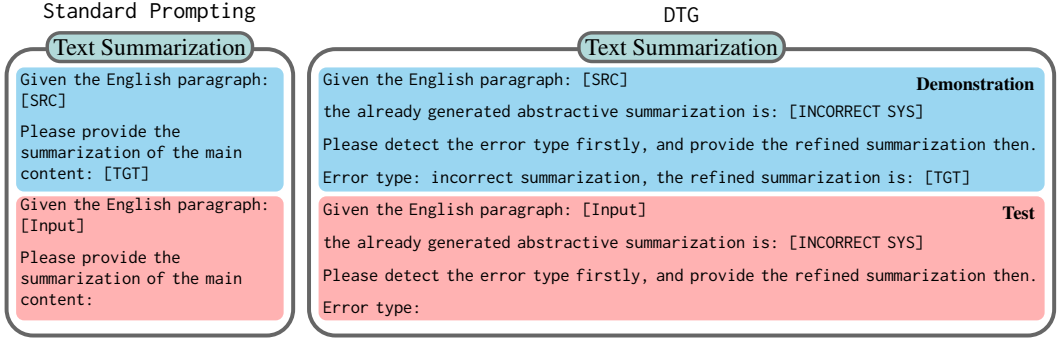
\begin{figure}
\centering
    
\begin{tikzpicture}

\scriptsize{
\begin{scope}[]

\node [anchor=south west] (n1) at (0, 0) {\small \texttt{Standard Prompting}};

\node [anchor=north,rectangle,rounded corners=10pt,minimum height=1.5in,minimum width=1.5in, draw, line width=1.5pt, draw=black!60!white] (b1) at ([xshift=0em, yshift=-1em]n1.south) {};

\node [anchor=center,rectangle,rounded corners=5pt,minimum height=1em,minimum width=2em,fill=teal!30, draw=black!60!white, line width=1.3pt] (l1) at ([xshift=0em, yshift=0em]b1.north) {\small{Text Summarization}};

\node [anchor=north west,rectangle,rounded corners=2pt,minimum height=1em,minimum width=2em, text width=1.5in] (n21) at ([xshift=0.5em, yshift=-1em]b1.north west) {\texttt{{Given the English paragraph: [SRC]}}};
\node [anchor=north west,rectangle,rounded corners=2pt,minimum height=1em,minimum width=2em, text width=1.5in] (n22) at ([xshift=0em, yshift=-0.2em]n21.south west) {\texttt{{Please provide the summarization of the} {main content: [TGT]}}};
\node [anchor=north west,rectangle,rounded corners=2pt,minimum height=1em,minimum width=2em, text width=1.5in] (n23) at ([xshift=0em, yshift=-0.2em]n22.south west) {\texttt{{Given the English paragraph: [Input]}}};
\node [anchor=north west,rectangle,rounded corners=2pt,minimum height=1em,minimum width=2em, text width=1.5in] (n24) at ([xshift=0em, yshift=-0.2em]n23.south west) {\texttt{{Please provide the summarization of the} {main content:}}};

\begin{pgfonlayer}{background}
\node [anchor=north,rectangle,rounded corners=3pt,minimum height=0.65in,minimum width=1.4in,fill=WindowsBlue!40] (bb) at ([xshift=0em, yshift=-0.2em]l1.south) {};
\node [anchor=north,rectangle,rounded corners=3pt,minimum height=0.65in,minimum width=1.4in,fill=red!30] (br) at ([xshift=0em, yshift=-0.2em]bb.south) {};
\end{pgfonlayer}

\end{scope}
}

\scriptsize{
\begin{scope}[xshift=3.25in]

\tikzstyle{node1}=[rectangle,rounded corners=2pt,inner sep=0mm,minimum height=1.2em,minimum width=6em,fill=teal!45,font=\footnotesize]

\node [anchor=south west] (n1) at (0, 0) {\small \texttt{DTG}};

\node [anchor=north,rectangle,rounded corners=10pt,minimum height=1.5in,minimum width=3.9in, draw, line width=1.5pt, draw=black!60!white] (b1) at ([xshift=0em, yshift=-1em]n1.south) {};

\node [anchor=center,rectangle,rounded corners=5pt,minimum height=1em,minimum width=2em,fill=teal!30, draw=black!60!white, line width=1.3pt] (l1) at ([xshift=0em, yshift=0em]b1.north) {\small{Text Summarization}};

\node [anchor=north west,rectangle,rounded corners=2pt,minimum height=1em,minimum width=2em, text width=3.9in] (n21) at ([xshift=0.5em, yshift=-1em]b1.north west) {\texttt{{Given the English paragraph: [SRC]}}};
\node [anchor=north west,rectangle,rounded corners=2pt,minimum height=1em,minimum width=2em, text width=3.9in] (n22) at ([xshift=0em, yshift=-0.2em]n21.south west) {\texttt{{the already  generated abstractive summarization is: [INCORRECT SYS]}}};
\node [anchor=north west,rectangle,rounded corners=2pt,minimum height=1em,minimum width=2em, text width=3.9in] (n23) at ([xshift=0em, yshift=-0.2em]n22.south west) {\texttt{{Please detect the error type firstly, and provide the} {refined summarization then.}}};

\node [anchor=north west,rectangle,rounded corners=2pt,minimum height=1em,minimum width=2em, text width=3.9in] (n24) at ([xshift=0em, yshift=-0.2em]n23.south west) {\texttt{{Error type: incorrect summarization, the refined} {summarization is: [TGT]}}};

\node [anchor=north west,rectangle,rounded corners=2pt,minimum height=1em,minimum width=2em, text width=3.4in] (n25) at ([xshift=0em, yshift=-0.2em]n24.south west) {\texttt{{Given the English paragraph: [Input]}}};
\node [anchor=north west,rectangle,rounded corners=2pt,minimum height=1em,minimum width=2em, text width=3.9in] (n26) at ([xshift=0em, yshift=-0.2em]n25.south west) {\texttt{{the already  generated abstractive summarization is: [INCORRECT SYS]}}};
\node [anchor=north west,rectangle,rounded corners=2pt,minimum height=1em,minimum width=2em, text width=3.9in] (n27) at ([xshift=0em, yshift=-0.2em]n26.south west) {\texttt{{Please detect the error type firstly, and provide the} {refined summarization then.}}};
\node [anchor=north west,rectangle,rounded corners=2pt,minimum height=1em,minimum width=2em, text width=3.9in] (n28) at ([xshift=0em, yshift=-0.2em]n27.south west) {\texttt{{Error type:}}};

\begin{pgfonlayer}{background}
\node [anchor=north,rectangle,rounded corners=3pt,minimum height=0.65in,minimum width=3.8in,fill=WindowsBlue!40] (bb) at ([xshift=0em, yshift=-0.2em]l1.south) {};
\node [anchor=north east,fill=WindowsBlue!40] (bl) at ([xshift=-0.2em, yshift=-0.2em]bb.north east) {\color{black}\textbf{Demonstration}};
\node [anchor=north,rectangle,rounded corners=3pt,minimum height=0.65in,minimum width=3.8in,fill=red!30] (br) at ([xshift=0em, yshift=-0.2em]bb.south) {};
\node [anchor=north east,fill=red!30] (bl) at ([xshift=-0.2em, yshift=-0.2em]br.north east) {\color{black}\textbf{Test}};
\end{pgfonlayer}

\end{scope}
}
\end{tikzpicture}

\caption{Comparison of standard GPT prompting and our DTG prompt desgin for summarization task. Note that prompt in {blue} denotes the demonstration, and that in {red} denotes the test input. [SRC] and [Input] means the source input, {TGT} means the target reference and [INCORRECT SYS] means the irrelevant system output (e.g., such as a randomly sampled text or even an empty string).}
\label{fig:prompt_comparison}
\end{figure}

The main contributions of this work are summarized as follows:
\begin{itemize}[leftmargin=*]
    \item We propose a novel prompting framework named \textit{Deliberate then Generate}~(DTG) for LLMs. Extensive ablation studies and analyses show that by prompting the model to detect errors and refine, LLMs indeed deliberate and avoid possible errors in generation.
    \item We conduct experiments on 20+ datasets across 7 text generation tasks, where DTG prompting brings consistent improvements and achieves SoTA performance on several benchmarks.
    \item To the best of our knowledge, we are the first to evaluate the performance of GPT3.5 and GPT4 on multiple benchmark text generation tasks including text summarization, dialogue summarization, simplification, style transfer, paraphrase and commonsense generation. We hope the experimental results help deepen our understanding of SoTA LLMs.
\end{itemize}

\section{Related Work}

\paragraph{Large Language Models.}
With the scaling of model and corpus sizes, Large Language Models~(LLMs)~\citep{devlin2018bert,radford2019language,lewis2019bart} have achieved remarkable success in various areas of natural language processing. Considering the large scale of the LLMs, exploring cost-effective fine-tuning methods is one appealing line of work when adapting to downstream tasks~\citep{hu2021lora,li2021prefix}. The fine-tuning approach poses a challenge when applied to powerful black-box LLMs that only offer APIs for interaction, as it requires access to the underlying parameters. With the help of instruction tuning~\citep{wei2021finetuned} and reinforcement learning from human feedback~\citep{ouyang2022training}, recent LLMs can achieve gradient-free adaptation to various downstream tasks by prompting with natural language instructions, and some powerful capacities such as in-context learning~\citep{brown2020language} have also emerged.

\paragraph{Prompting Methods.} 
Prompting is a general method for humans to interact with LLMs, which is usually designed as an instruction for a task that guides LLMs toward intended outputs~\citep{schick2020exploiting,sanh2021multitask}. To make the most of LLMs on downstream tasks, the prompts need to be carefully designed, either manually~\cite{hendy2023good} or automatically~\citep{gao2020making,zhou2022large}. Prompting also provides a way to interact with LLMs in natural language, such as letting them utilize external tools~\citep{schick2023toolformer}, resources~\cite{ghazvininejad2023dictionary} and models~\citep{wu2023visual,shen2023hugginggpt}, or conducting Chain-of-Thought~(CoT) reasoning in generation~\citep{wei2022chain,kojima2022large}. A concurrent work incorporates answers in previous rounds into prompts in an iterative process to improve the accuracy of LLMs on reasoning tasks~\cite{zheng2023progressive}. Besides multi-step reasoning, basic prompts are still widely utilized in general text generation tasks such as machine translation and summarization, where previous advanced methods such as CoT have been shown ineffective~\citep{peng2023towards}.
In this paper, we propose Deliberate then Generate~(DTG), a simple and general prompting method that consistently improves model performance across various text generation tasks, without task-specific designs. 

\section{Deliberate then Generate}
\label{sec:method}
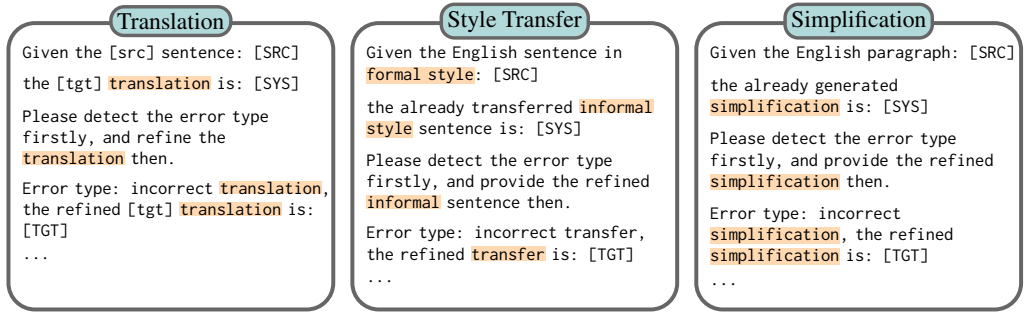
\begin{figure*}
    \centering

    \scriptsize{
\begin{tikzpicture}

\begin{scope}[]

\node [anchor=south west] (n1) at (0, 0) {};

\node [anchor=north,rectangle,rounded corners=10pt,minimum height=1.5in,minimum width=1.7in, draw, line width=1.5pt, draw=black!60!white] (b1) at ([xshift=0em, yshift=-1em]n1.south) {};

\node [anchor=center,rectangle,rounded corners=5pt,minimum height=1em,minimum width=2em,fill=teal!30, draw=black!60!white, line width=1.3pt] (l1) at ([xshift=0em, yshift=0em]b1.north) {\small{Translation}};

\node [anchor=north west,rectangle,rounded corners=2pt,minimum height=1em,minimum width=2em, text width=1.8in] (n21) at ([xshift=0.5em, yshift=-1em]b1.north west) {\texttt{Given the [src] sentence: [SRC]}};
\node [anchor=north west,rectangle,rounded corners=2pt,minimum height=1em,minimum width=2em, text width=1.6in] (n22) at ([xshift=0em, yshift=-0.2em]n21.south west) {\texttt{the [tgt] \colorbox{orange!30}{translation} is: [SYS]}};
\node [anchor=north west,rectangle,rounded corners=2pt,minimum height=1em,minimum width=2em, text width=1.6in] (n23) at ([xshift=0em, yshift=-0.2em]n22.south west) {\texttt{Please detect the error type firstly, and refine the \colorbox{orange!30}{translation} then.}};

\node [anchor=north west,rectangle,rounded corners=2pt,minimum height=1em,minimum width=2em, text width=1.6in] (n24) at ([xshift=0em, yshift=-0.2em]n23.south west) {\texttt{Error type: incorrect \colorbox{orange!30}{translation}, the refined [tgt] \colorbox{orange!30}{translation} is: [TGT]}};
\node [anchor=north west,rectangle,rounded corners=2pt,minimum height=1em,minimum width=2em, text width=1.6in] (n25) at ([xshift=0em, yshift=-0.2em]n24.south west) {\texttt{...}};

\end{scope}

\begin{scope}[xshift=1.8in]

\tikzstyle{node1}=[rectangle,rounded corners=2pt,inner sep=0mm,minimum height=1.2em,minimum width=6em,fill=teal!45,font=\footnotesize]

\node [anchor=south west] (n1) at (0, 0) {};

\node [anchor=north,rectangle,rounded corners=10pt,minimum height=1.5in,minimum width=1.7in, draw, line width=1.5pt, draw=black!60!white] (b1) at ([xshift=0em, yshift=-1em]n1.south) {};

\node [anchor=center,rectangle,rounded corners=5pt,minimum height=1em,minimum width=2em,fill=teal!30, draw=black!60!white, line width=1.3pt] (l1) at ([xshift=0em, yshift=0em]b1.north) {\small{Style Transfer}};

\node [anchor=north west,rectangle,rounded corners=2pt,minimum height=1em,minimum width=2em, text width=1.6in] (n21) at ([xshift=0.5em, yshift=-1em]b1.north west) {\texttt{Given the English sentence in \colorbox{orange!30}{formal style}: [SRC]}};
\node [anchor=north west,rectangle,rounded corners=2pt,minimum height=1em,minimum width=2em, text width=1.6in] (n22) at ([xshift=0em, yshift=-0.2em]n21.south west) {\texttt{the already transferred \colorbox {orange!30}{informal} \colorbox {orange!30}{style} sentence is: [SYS]}};
\node [anchor=north west,rectangle,rounded corners=2pt,minimum height=1em,minimum width=2em, text width=1.6in] (n23) at ([xshift=0em, yshift=-0.2em]n22.south west) {\texttt{Please detect the error type firstly, and provide the refined \colorbox {orange!30}{informal} sentence then.}};

\node [anchor=north west,rectangle,rounded corners=2pt,minimum height=1em,minimum width=2em, text width=1.6in] (n24) at ([xshift=0em, yshift=-0.2em]n23.south west) {\texttt{Error type: incorrect transfer, the refined \colorbox {orange!30}{transfer} is: [TGT]}};
\node [anchor=north west,rectangle,rounded corners=2pt,minimum height=1em,minimum width=2em, text width=1.6in] (n25) at ([xshift=0em, yshift=-0.2em]n24.south west) {\texttt{...}};
\end{scope}

\begin{scope}[xshift=3.6in]

\node [anchor=south west] (n1) at (0, 0) {};

\node [anchor=north,rectangle,rounded corners=10pt,minimum height=1.5in,minimum width=1.7in, draw, line width=1.5pt, draw=black!60!white] (b1) at ([xshift=0em, yshift=-1em]n1.south) {};

\node [anchor=center,rectangle,rounded corners=5pt,minimum height=1em,minimum width=2em,fill=teal!30, draw=black!60!white, line width=1.3pt] (l1) at ([xshift=0em, yshift=0em]b1.north) {\small{Simplification}};

\node [anchor=north west,rectangle,rounded corners=2pt,minimum height=1em,minimum width=2em, text width=1.6in] (n21) at ([xshift=0.5em, yshift=-1em]b1.north west) {\texttt{Given the English paragraph: [SRC]}};
\node [anchor=north west,rectangle,rounded corners=2pt,minimum height=1em,minimum width=2em, text width=1.6in] (n22) at ([xshift=0em, yshift=-0.2em]n21.south west) {\texttt{the already generated \colorbox{orange!30}{simplification} is: [SYS]}};
\node [anchor=north west,rectangle,rounded corners=2pt,minimum height=1em,minimum width=2em, text width=1.6in] (n23) at ([xshift=0em, yshift=-0.2em]n22.south west) {\texttt{Please detect the error type firstly, and provide the refined \colorbox{orange!30}{simplification} then.}};

\node [anchor=north west,rectangle,rounded corners=2pt,minimum height=1em,minimum width=2em, text width=1.6in] (n24) at ([xshift=0em, yshift=-0.2em]n23.south west) {\texttt{Error type: incorrect \colorbox{orange!30}{simplification}, the refined \colorbox{orange!30}{simplification} is: [TGT]}};
\node [anchor=north west,rectangle,rounded corners=2pt,minimum height=1em,minimum width=2em, text width=1.6in] (n25) at ([xshift=0em, yshift=-0.2em]n24.south west) {\texttt{...}};

\end{scope}

\end{tikzpicture}
}
\caption{Illustration of DTG demonstration design for machine translation, style transfer and text simplification tasks. Due to the limited page, please refer to the Appendix for the remained 3 generation tasks, including dialogue summarization, paraphrase and commonsense generation.}
\label{fig:DTG_demonstration}
\end{figure*}

Language acquisition by a human is normally based on both positive and negative feedback and improves the ability of language use through corrections. Inspired by this, unlike the conventional prompts only with correct information, we introduce a more deliberate approach termed Deliberate then Generate (DTG) prompting by facilitating LLMs to \emph{detect errors on a synthesized text that may contain errors}. Specifically, the proposed DTG method unfolds in the following manner: 1) It begins by a concise and explicit instruction of the desired task, providing guidance on generating an intended text based on a given input text; 2) A synthesized text is then provided as a candidate output; (3) Finally, DTG encourages the model to detect potential errors, and subsequently generate an improved output after thorough deliberation.

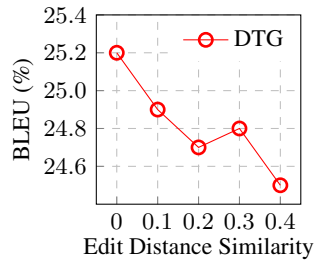
\begin{wrapfigure}[13]{r}{0.3\textwidth}
\pgfplotsset{}
\vspace{-0.5cm}
    \begin{centering}
    \begin{tikzpicture}
        \pgfplotsset{footnotesize,samples=2}
        \begin{groupplot}[
            group style = {group size = 2 by 1, horizontal sep = 55pt}, ]
            \nextgroupplot[
                align = center,
                width=4.3cm, height=4.1cm,
                xmin=-0.05, xmax=0.45,
                ymin=24.4, ymax=25.4,
                xtick={0.0, 0.1, 0.2, 0.3, 0.4},
                xlabel={Edit Distance Similarity},
                ylabel={BLEU (\%)},
                ylabel style={align=center},
                ytick={24.6, 24.8, 25.0, 25.2, 25.4},
                yticklabel style={/pgf/number format/precision=1,/pgf/number format/fixed zerofill},
                x label style={at={(axis description cs:0.5,0.08)},anchor=north},
                y label style={at={(axis description cs:0.19,0.5)},anchor=south},
                xtick pos=bottom,
                ytick pos=left,
                xmajorgrids,
                ymajorgrids,
                major grid style={dashed},
                legend style={anchor=north east,draw=none,xshift=-0cm,yshift=0.0cm},
                ]
                \addplot[
                    color=red,
                    mark=o,
                    mark size=2.5pt,
                    mark options={fill=red ,line width=1.0pt}
                    ]
                    coordinates {
                    (0,   25.2)
                    (0.1,   24.9)
                    (0.2,  24.7)
                    (0.3,  24.8)
                    (0.4,  24.5)
                    };
                \addlegendentry{DTG}
        \end{groupplot}
    \end{tikzpicture}
    \caption{
    BLEU scores against the similarity (Edit Distance) on ZH-EN task.}
    \vspace{-3mm}
    \label{fig:edit_distance}
    \end{centering}
\end{wrapfigure}
Figure~\ref{fig:prompt_comparison} illustrates a comparison between standard prompting and our proposed DTG prompting for the summarization task in the one-shot scenario. A distinctive feature of DTG is its emphasis on error detection other than immediate response. Instead of generating the outcome directly from the given input text, DTG steers the model to make deliberate decisions by detecting the error type firstly based on both the input text, denoted as ``[SRC]'', and a pre-defined candidate, denoted as ``[SYS]'', before the final decisions. This deliberative process forms the bedrock of the DTG approach and will be further elaborated upon in the analysis section (i.e., Section~\ref{sec:analysis}). Besides, a few demonstrations can be provided,
imbuing LLMs with an awareness of the expected output (highlighted in blue), and the test input (marked in red). DTG is a general prompting method that could be easily applied to any text generation task with minimal modifications to the prompt. Figure~\ref{fig:DTG_demonstration} illustrates the particular prompts used for 3 generation tasks we considered, indicating that minimal customization is required across different tasks as highlighted in yellow.

The determination of the synthesized text is another key part of DTG. Straightforwardly, using the output of a baseline system, which can either be LLMs themselves or any other models, is a natural choice. However, such baseline text just requires minor modifications, and thus cannot well trigger the deliberation ability of LLMs. 
Moreover, we find that the lower the similarity between the candidate and the reference, the better the quality of the generated text.
As shown in Figure~\ref{fig:edit_distance}, we select sentences that have various similarities with the reference~(using edit distance) as the synthesized sentence, and the performance decreases monotonically in general when the similarity increases.
Therefore, we seek to choose a sentence that does not contain any correct information as the synthesized text. Potential candidates include a randomly sampled sentence or more extremely, an \textit{empty string}, i.e., setting ``[SYS]'' as `` ''. Both choices successfully facilitate deliberation and consistently improve the outcomes across multiple text generation tasks. We use an empty string in our experiments as it is more general and elegant.

DTG has the following exceptional properties to steer LLMs on various text generation tasks:
\begin{itemize}[leftmargin=*]
    \item \emph{Simple}: The final results can be obtained through a single-step inference of the LLM, without any additional resources or costs.
    \item \emph{General}: It can be effortlessly applied to a broad range of text generation tasks only with minimal adjustments in the prompt.
\end{itemize}

\section{Datasets and Evaluation}
In experiments, we are devoted to evaluating the generation ability of LLMs and the proposed DTG prompting.
We select 7 representative generation tasks, including machine translation, abstractive summarization, dialogue summarization, text simplification, style transfer, paraphrase and commonsense generation.

\paragraph{Machine Translation}
For the machine translation task, we aligned with \citet{hendy2023good}'s work and experimented on both high-resource and low-resource scenarios. For the high-resource setting, we include German, Czech, Chinese, Japanese, Russian, and Ukrainian paired with English. In the low-resource context, we examine Icelandic and Hausa. The performance is evaluated in terms of SacreBLEU\footnote{BLEU+case.mixed+numrefs.1+smooth.exp+tok.13a+version.2.3.1}~\cite{post-2018-call}, ChrF, TER (translation error rate) and COMET-22~\cite{rei-etal-2022-comet}.

\paragraph{Abstractive Summarization}
We also evaluate LLM's ability to process long sequence on CNN-DailyMail and Gigaword, two widely used abstractive summarization datasets. The evaluation metric is F1-Rouge \cite{lin-2004-rouge}, consisting of Rouge-1, Rouge-2 and Rouge-L.

\paragraph{Dialog Summarization}
Dialogue summarization presents greater challenges than traditional text summarization due to the intricate conversation contexts that models need to comprehend, though their contexts are relatively shorter. This attribute enables us to test few-shot abilities due to the restricted input length. To investigate this, we select SamSum\footnote{https://huggingface.co/datasets/samsum} \cite{gliwa-etal-2019-samsum} and DialogSum\footnote{https://github.com/cylnlp/DialogSum} \cite{chen-etal-2021-dialogsum}, two benchmark datasets for dialogue summarization. The evaluation metric is the same as abstractive summarization.

\paragraph{Text Simplification}
The purpose of text simplification is to revise complex text into sequences with simplified grammar and word choice. In this work, we mainly report the performance on two benchmarks, namely Asset \cite{alva-manchego-etal-2020-asset} and Wiki-auto \cite{jiang-etal-2020-neural}. Asset is a multi-reference dataset for the evaluation of sentence simplification in English. The dataset uses the same 2,359 sentences from TurkCorpus \cite{Xu-EtAl:2016:TACL} and each sentence is associated with 10 crowdsourced simplifications. Similarly, each test set in Wiki-auto owns 8 references. We use SacreBLEU and BLEURT as the metric.

\begin{table}[!t]
\centering
\caption{Evaluation results of GPT and GPT4 on six high-resource and two-low resource machine translation tasks from WMT Testsets. The best scores across different systems are marked in bold.}
\label{tab:mt}
\resizebox{\textwidth}{!}{
\begin{tabular}{l|c c c c | c c c c}
\hline
\bf{System} &  \bf{COMET-22$\uparrow$}  & \bf{TER$\downarrow$} & \bf{ChrF$\uparrow$}  & \bf{BLEU$\uparrow$} &
\bf{COMET-22$\uparrow$}  & \bf{TER$\downarrow$} & \bf{ChrF$\uparrow$}  & \bf{BLEU$\uparrow$} \\ 
\hline
& \multicolumn{4}{c|}{DE-EN} &   \multicolumn{4}{c}{ZH-EN} \\ 

WMT-Best\dag            & 85.0     & 51.5   & 58.5     & 33.4      & 81.0     & \bf 54.7 & \bf 61.1 & \bf 33.5  \\
MS-Translator\dag       & 84.7     & \bf 51.2   &  58.5 & \bf{33.5} & 80.4     & 60.0 & 57.7     & 27.9      \\
GPT 1-shot              & 84.7     & 53.7   & 56.2     & 30.4      & 81.0     & 64.4 & 54.9     & 23.7      \\
\rowcolor{gray!40}
\; + DTG                & 85.0     & 52.4   & 57.7     & 32.3      & 81.4     & 63.6 & 56.2     & 25.3  \\
GPT 5-shot              & 85.3     & 52.3   & 57.6     & 32.3      & 81.1     & 63.7 & 54.6     & 23.6      \\
\rowcolor{gray!40} 
\; + DTG                & 85.4 & 51.9   & 58.2     & 33.2      & 81.7 &	62.4 & 55.9     & 25.2      \\
GPT4 1-shot             &   85.6        &   51.7     &   \bf 58.9      &    \bf 33.5       & 82.4     & 62.8     & 57.3 & 26.0 \\
\rowcolor{gray!40}
\; + DTG                & \bf 85.8    & 51.8   &   58.8   &   33.4    & \bf 82.9     & 62.0 & 57.3     & 25.7  \\
\hline

& \multicolumn{4}{c|}{CS-EN} &   \multicolumn{4}{c}{RU-EN} \\ 
WMT-Best\dag            & \bf 89.0 &	\bf{26.8} &	\bf{79.3} &	\bf{64.2} & 86.0 & 	\bf 43.8 &	\bf{68.9} & 	\bf{45.1} \\
MS-Translator\dag       & 87.4	   & 34.5&	74.0	& 54.9 & 85.2&	45.1&	68.3&	43.9  \\
GPT 1-shot              & 86.2     & 43.7 &	67.5 & 44.8  &	84.8 &	48.2 &	65.3 & 39.7 \\
\rowcolor{gray!40} 
\; + DTG                & 86.7     & 42.6 &	68.8 & 45.6  &	85.0 &	48.3 &	66.1 & 40.0 \\
GPT 5-shot              & 86.9     & 40.7 &	69.2 & 47.2  &	84.9 &	48.0 &	65.2 & 39.9 \\
\rowcolor{gray!40} 
\; + DTG                & 87.0     & 40.9 &	69.6 & 47.4  &	85.1 &	47.7 &	66.2 &  40.3 \\
GPT4 1-shot             & 87.3         &   40.3     &    70.9     &    48.1       & 86.1     & 45.2     & 68.5 & 43.1 \\
\rowcolor{gray!40}
\; + DTG                &  87.3   &  39.8  &  70.9    &   48.9    & \bf 86.3     & 45.1 & 68.5     & 43.1  \\
\hline

& \multicolumn{4}{c|}{JA-EN} &   \multicolumn{4}{c}{UK-EN} \\ 
WMT-Best\dag            & 81.6     & 69.4 &	49.8 &	\bf{24.8} & 	\bf{86.0}	& \bf 42.7 &	\bf{67.3} &	 \bf{44.6} \\
MS-Translator\dag       & 81.5	   & \bf 69.0 & 49.6 &	24.5 & 83.5&	45.7	&65.3&	42.4 \\ 

GPT 1-shot              & 81.3     & 74.4 &  47.9 & 21.5  &	83.5 &	50.5 &	61.1 & 36.8 \\
\rowcolor{gray!40} 
\; + DTG                & 81.7     & 74.6 &	47.9 & 21.4  &	84.0 &	49.9 &	61.7 & 37.1 \\
GPT 5-shot              & 81.2     & 74.2 &	47.0 & 20.5  &	84.0 &	49.2 &	61.9 & 38.0 \\
\rowcolor{gray!40} 
\; + DTG                &  82.2 &	72.6 &	48.2 & 22.4  &	84.2 &	48.4 &	62.6 & 39.0 \\
GPT4 1-shot             &  83.4        &  69.6      &   \bf 51.1       &   24.7        & 85.7     & 46.9     & 65.2 & 39.9 \\
\rowcolor{gray!40}
\; + DTG                & \bf 83.6   &  69.5  &  \bf 51.1    & \bf 24.8      & 85.7     & 47.1 & 65.2     & 39.9  \\
\hline

& \multicolumn{4}{c|}{IS-EN} &   \multicolumn{4}{c}{HA-EN} \\
WMT-Best\dag            & \bf 87.0 & \bf 44.8	& 62.3  &	\bf{41.7} & \bf{80.0} &	\bf 69.0 &	\bf{48.7} &	\bf{21.0} \\
MS-Translator\dag       & 85.9     & 45.2&	62.8 &	40.5& 73.3&	73.4&	43.4&	16.2\\

GPT 1-shot              & 83.5     & 52.7 &	57.0 & 33.6  &	78.0 &	72.8 &	47.3 & 18.6 \\
\rowcolor{gray!40} 
\; + DTG                & 84.0     & 51.7 &	58.3 & 35.2  &	78.3 &	74.8 &	48.0 & 18.6 \\
GPT 5-shot              & 84.1     & 50.6 &	58.0 & 35.0  &	78.3 &	72.2 &	47.6 & 18.8 \\
\rowcolor{gray!40} 
\; + DTG                & 84.6     & 50.2 &	58.8 &  36.0  &	78.6 &	71.9 &	48.0 &  19.2 \\
GPT4 1-shot             &  86.9        &  47.0      &   63.8       &   39.9        & 77.5     & 75.7     & 47.8 & 18.3 \\
\rowcolor{gray!40}
\; + DTG                &  \bf 87.0   &  46.7  & \bf 63.9    & 40.3      & 77.9     & 75.1 & 47.9     & 18.7  \\
\hline

\end{tabular}}
\end{table}

\paragraph{Style Transfer}
We used three widely-used English transfer learning datasets, namely Grammalry's Yahoo Answers Formality Corpus (GYAFC), Amazon and Yelp reviews. 
The GYAFC dataset \cite{rao-tetreault-2018-dear} was originally a question-and-answer dataset on an online forum, consisting of informal and formal sentences from the two categories: Entertainment \& Music (EM) and Family \& Relationships (FR). Both FR and EM provide 4 references to evaluate the fidelity. The Amazon dataset is a product review dataset, labeled as either a positive or negative sentiment. Similarly, the Yelp dataset is a restaurant and business review dataset with positive and negative sentiments. Both Amazon and Yelp are single-reference. The evaluation metrics contain BLEU and BLEURT~\citep{sellam2020bleurt}.

\paragraph{Paraphrase}
We endeavor to evaluate the paraphrase ability of LLMs upon the well-known Quora Question Pairs (QQP) dataset, which requires generating an alternative surface form in the same language expressing the same semantic content. We utilize the preprocessed data from \citep{gong2022diffuseq}.

\paragraph{Common Sense Generation}
We choose CommonGen~\cite{lin-etal-2020-commongen}, a novel constrained generation task that requires models to generate a coherent sentence with the providing key concepts.

We summarize the details of each dataset for each task, including the test sets, the selection of demonstrations (mostly from validation sets) and the corresponding prompts we have used. For more details please refer to the attached Appendix.

\section{Experiments}
In this section, we assess the efficacy of the \texttt{text-davinci-003} (also known as GPT3.5, which is denoted as GPT in the following for simplicity) across 7 sequence generation tasks, including machine translation, abstractive summarization, dialogue summarization, text simplification, style transfer, commonsense generation and paraphrase. 
The chosen baseline comparisons consist of 1-shot, and few-shot (mostly 5-shot) learning scenarios. It is worth mentioning that while the performance of GPT models on machine translation has been extensively investigated in previous research, other generation tasks (e.g., text simplification and style transfer) have received comparatively limited attention. To demonstrate the versatility of DTG method and address the primary limitation of GPT3.5, we conduct further experiments with GPT4, a cutting-edge LLM API. Due to the considerable computational cost and API request constraints associated with the GPT4, it is challenging to perform extensive experiments. In the current manuscript, we only report the results on machine translation and text simplification. We aim to highlight the significant potential of GPT models to excel in downstream tasks without the necessity for fine-tuning.

\begin{table*}[!t]
\centering
\caption{Experimental results on four summarization tasks.}
\setlength\tabcolsep{2.5pt}
\small
{
\begin{tabular}{l|ccc|ccc|ccc|ccc}
\hline
\multirow{2}{*}{\textbf{System}}&  \multicolumn{3}{c|}{\textbf{CNN/DailyMail}} & \multicolumn{3}{c|}{\textbf{GigaWord}} & \multicolumn{3}{c|}{\textbf{SamSum}} & \multicolumn{3}{c}{\textbf{DialogSum}}\\  
& R1 & R2 & RL & R1 & R2 & RL & R1 & R2 & RL & R1 & R2 & RL\\
\hline
Transformer \cite{vaswani2017attention}         & 40.47 & 17.73 & 37.29 & 37.57 & 18.90 & 34.69  & 37.20 & 10.86 & 34.69 & 35.91 & 8.74  & 33.50 \\
BART \cite{lewis-etal-2020-bart}                & \bf 44.16 & \bf 21.28 & \bf 40.90 & \bf 39.29 & \bf 20.09 & \bf 35.65  & \bf 53.12 & \bf 27.95 & \bf 49.15 & \bf 47.28 & \bf 21.18 & 44.83 \\
UniLMv2 \cite{bao2020UniLMv2}                   & 43.16 & 20.42 & 40.14 & -     & -     & -      & 50.53 & 26.62 & 48.81 & 47.04 & 21.13 & \bf 45.04 \\
GPT 1-shot                                      & 38.87 & 15.36 & 35.11 & 31.24 & 11.61 & 27.99  & 44.52 & 19.92 & 39.60 & 36.84 & 14.23 & 32.20 \\
\rowcolor{gray!40} 
\; + DTG                                        & 40.17 & 15.60 & 36.04 & 31.50 & 12.00 & 28.50  & 45.50 & 20.58 & 40.13 & 39.01 & 15.50 & 34.13 \\
GPT 5-shot                                      & -     & -     & -     & 33.04 & 12.78 & 29.86  & 46.44 & 20.69 & 41.10 & 40.86 & 17.10 & 35.78 \\
\rowcolor{gray!40}
\; + DTG                                        & -     & -     & -     & 33.54 & 13.63 & 30.36  & 48.72 & 23.16 & 43.23 & 42.64 & 18.12 & 37.57 \\
GPT 10-shot                                      & -     & -     & -     & 33.24 & 13.26 & 30.46  & 47.37 & 22.08 & 42.20 & 41.28 & 17.48 & 36.69 \\
\rowcolor{gray!40}
\; + DTG                                        & -     & -     & -     & 34.02 & 14.21 & 31.04  & 50.48 & 24.88 & 45.31 & 45.11 & 19.50 & 39.71\\ 
\hline
\end{tabular}
}
\label{tab:summarization}
\end{table*} 

\subsection{Results on Machine Translation}
We compare the performance of GPT standard prompting and our deliberate then generate method (DTG) with that of a commercial system (Microsoft Translator) in addition to WMT SoTA systems. Table \ref{tab:mt} presents the results in both 1-shot and 5-shot scenarios. 
Without meticulous parameter tuning, we set the \textit{temperature} to 0 and \textit{top\_p} to 1 when calling the API. The findings here indicate that our re-implementation aligns with the trends observed in previous study~\citep{hendy2023good}, that 5-shot beats 1-shot in most language pairs. Benefiting from the deliberation, DTG effectively pushes the boundaries and leads to enhanced results across all to-English language pairs in both 1-shot and 5-shot settings based on GPT3.5 model. For instance, DTG method exhibits substantial BLEU score increases in DE-EN, ZH-EN, and UK-EN language pairs in 5 shot scenarios. More concretely, DTG even beats WMT-Best system in terms of COMET-22, which is a more recognized metric recently in the machine translation literature. Moreover, the consistent improvements on IS-EN and HA-EN demonstrate the effectiveness of DTG in low-resource settings.

We only conduct experiments on 1-shot scenario due to the limited access, and leave the remained 5-shot explorations as future work. We observe GPT4 1-shot can beat GPT3.5 5-shot by a large margin in most language pairs. Meanwhile, DTG is still effective on GPT4. This finding demonstrates much stronger LLMs can still benefit from deliberation.

\subsection{Results on Summarization}
For abstractive summarization, we mainly evaluate GPT models on CNN/DailyMail and GigaWord, two of the most widely-used summarization tasks. Due to the limit of max length for GPT models (4097) and the long input length of CNN/DailyMail, we only evaluate the performance in 1-shot scenario. As shown in Table \ref{tab:summarization}, GPT models show comparative performance with Transformer which is specially tuned on the downstream training set. Our DTG can also shown further improvement in terms of three Rouge metrics, which demonstrate the effectiveness of DTG on long-term modeling task. However, DTG still falls lag behind of large-scale pretrained models, such as BART~\cite{lewis-etal-2020-bart} and UniLMv2~\cite{bao2020UniLMv2} in automatic evaluations. We will add more human alignment judgment in Section \ref{sec:analysis}.

Dialogue generation represents a critical aspect of language tasks. In this context, we further corroborate the efficacy of Large Language Models (LLMs) in dialogue summarization, a composite task encapsulating elements of both dialogue and summarization. It is important to note that the results for DialogSum are averaged over three individual scores, each calculated using unique references spanning a range of topics. As observed, GPT 1-shot achieves commendable results compared to constrained systems, e.g., Transformer. Furthermore, DTG substantially incites GPT models to generate more precise summaries derived from extensive multi-turn dialogues. An upward trend in performance is observed with the introduction of additional demonstrations, further underscoring the effectiveness of the DTG method. Nonetheless, in the absence of specialized fine-tuning, the GPT3.5 model falls short of surpassing the performance of BART. Despite this, the model's performance remains notably impressive, highlighting the potential of LLMs in complex language generation tasks.

\subsection{Results on Style Transfer}
Table \ref{tab:style_transfer} displays performance across style transfer tasks from the GYAFC dataset: Entertainment Music (EM) and Family Relationships (FR), both involving informal to formal transformations. Evidently, the Deliberate then Generate (DTG) method prompts the GPT model to correct inaccuracies and generate more precise informal sentences. Specifically, DTG achieves an 8-point and 10.04-point increase in BLEU score for EM and FR tasks, respectively, compared to standard prompting. Although DTG trails BART \cite{lewis-etal-2020-bart} in BLEU scores, it surpasses BART in BLEURT scores, registering gains of 0.98 and 0.32 for EM and FR tasks, respectively. These results highlight the potential of LLMs and our DTG method in style transfer tasks.

\subsection{Results on Text Simplification}
\begin{wraptable}[13]{r}{0.58\textwidth}
\vspace{-0.20in}
\small
\caption{Comparisons of 1-shot, 5-shot with and without our DTG method on two text simplification tasks.}
\centering
\begin{tabular}{l|cc|cc}
\hline
\multirow{2}{*}{\textbf{System}} & \multicolumn{2}{c|}{\textbf{Asset}}  & \multicolumn{2}{c}{\textbf{Wiki-auto}} \\
&  \textbf{BLEU}  & \textbf{SARI} &  \textbf{BLEU}  & \textbf{SARI}\\
\hline
MUSS \citep{martin-etal-2022-muss}              & 72.9      & 44.15        & -         & 42.59      \\
Control Prefix \citep{clive-etal-2022-control}  & -         & 43.58        & -         & -          \\
TST-Final \citep{omelianchuk-etal-2021-text}    &-          & 41.46        & -         & -          \\
GPT 1-shot                                      & 67.6      & 46.12        & 65.0     & 44.97      \\  
\rowcolor{gray!40}
\; + DTG                                        & 72.9      & 47.23        & 72.0     & 47.15      \\
GPT 5-shot                                      & 73.3      & 45.95        & 70.0     & 45.12      \\
\rowcolor{gray!40}
\; + DTG                                        & \bf 80.2  & 47.05        & \bf 80.0 & \bf 47.54  \\
GPT4 5-shot                                     & 68.0      & 47.10        & 65.1     & 45.96      \\
\rowcolor{gray!40}
\; + DTG                                        & 74.9      & \bf 47.67    & 67.9     & 47.03      \\
\hline
\end{tabular}
\label{tab:simplification}
\end{wraptable}
Experiments were conducted on two text simplification benchmarks, Asset and Wiki-Auto, where the primary goal of which is to create a simplified rendition of the given text input. The main evaluation metric is the SARI score. Our findings illustrate that GPT models demonstrate robust performance across both simplification benchmarks, even surpassing the existing state-of-the-art models (MUSS) built based on BART. Furthermore, the incorporation of DTG method significantly enhances GPT model performance, leading to improvements in both BLEU and SARI scores. Specifically, DTG establishes a new benchmark for state-of-the-art results on these two simplification tasks. 

We also observe similar competitiveness of DTG on the two other style transfer benchmarks. It outperforms the standard prompting method with identical configurations in terms of both BLEU and BLEURT scores, further attesting to its efficacy. Again, GPT4 is superior to GPT3.5 and DTG also works at this time, though the obtained improvement is slightly marginal than that of GPT3.5.

\begin{table}[!t]
    \small
    \caption{Comparisons of 1-shot and 5-shot on four style transfer tasks, including Entertainment Music, Family Relationships, Amazon and Yelp. \dag denotes results borrowed from \citep{lai-etal-2021-thank}.}
    \setlength\tabcolsep{3.0pt}
    \centering
    \begin{tabular}{l|cc|cc|cc|cc}
    \hline
    \multirow{2}{*}{\textbf{System}} & \multicolumn{2}{c|}{\textbf{GYAFC \& EM}}  & \multicolumn{2}{c|}{\textbf{GYAFC \& FR}} & \multicolumn{2}{c|}{\textbf{Amazon}}  & \multicolumn{2}{c}{\textbf{Yelp}} \\
    &  \textbf{BLEU}  & \textbf{BLEURT}  &  \textbf{BLEU}  & \textbf{BLEURT} &  \textbf{BLEU}  & \textbf{BLEURT} &  \textbf{BLEU}  & \textbf{BLEURT}\\
    \hline
    Transformer\dag \cite{vaswani2017attention}     & 40.3     & -          & 47.7     & -          & -     & -    & -     & -      \\
    BART\dag \cite{lewis-etal-2020-bart}            & \bf 76.9 & 75.38      & \bf 79.3 & 75.11      & -     & -    & -     & -      \\
  
    GPT 1-shot                                      & 52.9     & 73.42      & 44.6     & 70.73      & 36.1     & 64.56    & 30.9     & 64.03      \\   
    \rowcolor{gray!40}
    \; + DTG                                        & 66.8     & 75.20      & 65.9     & 74.60      & 35.4     & 63.60    & 31.3     & 64.19      \\
    GPT 5-shot                                      & 61.3     & 75.40      & 63.9     & 74.35      & 39.3     & 64.76    & 31.4     & 64.16      \\
    \rowcolor{gray!40}
    \; + DTG                                        & 69.9     & \bf 76.36  & 74.1     & \bf 75.43      & \bf 40.9     & \bf 65.42    & \bf 32.2     & \bf 64.87      \\
    \hline
    \end{tabular}
    \label{tab:style_transfer}
    \end{table}

\paragraph{Results on Commonsense Generation}

\begin{wraptable}[9]{r}{0.38\textwidth}
\caption{Results on the CommonGen benchmark.}
\centering
\small
\setlength{\tabcolsep}{1.5pt}
\begin{tabular}{lcc}
\toprule
\bf Model & \bf BLEU-3/4 &\bf Rouge-2/L \\
\midrule
BART \cite{lewis-etal-2020-bart}    & 36.3/26.4 & 22.23/41.98  \\
T5-Large \cite{Colin2020T5}         & 39.0/28.6 & 22.01/42.97  \\
GPT 5-shot                           & 39.7/30.0 & 25.28/46.55  \\
\rowcolor{gray!40}
\; + DTG                            & \bf 43.2/33.5 & \bf 27.02/48.47  \\
\bottomrule
\end{tabular}

\label{tab:commongen}
\end{wraptable}
Table \ref{tab:commongen} summarizes the comparison between GPT models with and without DTG method on an open Commonsense generation benchmark. This task is more flexible than the aforementioned, meanwhile raising the evaluation difficulty. We see that GPT models with standard prompting even surpasses large-scale pretrained generation models, such as BART~\cite{lewis2019bart} and T5~\cite{Colin2020T5}. Our DTG achieves further improvements in terms of BLEU-3/BLEU-4 and Rouge-2/Rouge-L, resulting in an average of 3.50 BLEU scores and almost 2.00 Rouge score improvements. This also establishes a new SoTA on this benchmark.

\paragraph{Results on Paraphrase}

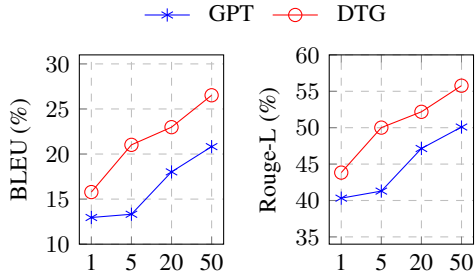
\begin{wrapfigure}[12]{r}{0.45\textwidth}
\pgfplotsset{}
\vspace{-0.65cm}
    \begin{centering}
    \begin{tikzpicture}
        \pgfplotsset{footnotesize,samples=2}
        \begin{groupplot}[
            group style = {group size = 2 by 1, horizontal sep = 40pt}, ]
            \nextgroupplot[
                align = center,
                legend style={at={(1.25,1.5)},anchor=north},
                width=3.5cm, height=4.1cm,
                ymin=10, ymax=31,
                symbolic x coords={1, 5, 20, 50},
                xtick=data,
                legend style={draw=none},
                ylabel={BLEU (\%)},
                ylabel style={align=center},
                ytick={10, 15, 20, 25, 30},
                x label style={at={(axis description cs:0.5,-0.15)},anchor=north},
                y label style={at={(axis description cs:0.37,0.5)},anchor=south},
                xtick pos=bottom,
                ytick pos=left,
                xmajorgrids,
                ymajorgrids,
                major grid style={dashed},
                ]
                \addplot[
                    color=blue,
                    mark=asterisk,
                    mark size=2.5pt,
                    ]
                    coordinates {
                    (1,   12.96)
                    (5,   13.32)
                    (20,  18.02)
                    (50,  20.82)
                    };
                \addplot[
                    color=red,
                    mark=o,
                    mark size=2.5pt,
                    ]
                    coordinates {
                    (1,   15.78)
                    (5,   21.02)
                    (20,  22.98)
                    (50,  26.52)
                    };
            \nextgroupplot[
                align = center,
                width=3.5cm, height=4.1cm,
                ymin=34, ymax=60,
                symbolic x coords={1, 5, 20, 50},
                xtick=data,
                ylabel={Rouge-L (\%)},
                ylabel style={align=center},
                ytick={35, 40, 45, 50, 55, 60},
                x label style={at={(axis description cs:0.07,-0.22)},anchor=north},
                y label style={at={(axis description cs:0.37,0.5)},anchor=south},
                legend style={anchor=south,draw=none,xshift=-2.7cm,yshift=0.3cm,column sep=5pt},
                ytick pos=left,
                xmajorgrids,
                ymajorgrids,
                major grid style={dashed},
                legend columns = 3,
                ]
                \addplot[
                    color=blue,
                    mark=asterisk,
                    mark size=2.5pt,
                    bar shift =0.1cm
                    ]
                    coordinates {
                    (1,   40.32)
                    (5,   41.29)
                    (20,  47.14)
                    (50,  50.09)
                    };
                    \addlegendentry{GPT}
                \addplot[
                    color=red,
                    mark=o,
                    mark size=2.5pt,
                    bar shift =0.1cm
                    ]
                    coordinates {
                    (1,   43.83)
                    (5,   50.01)
                    (20,  52.18)
                    (50,  55.77)
                    };
                    \addlegendentry{DTG}
            
        \end{groupplot}
    \end{tikzpicture}
    \caption{
    BLEU and Rouge-L scores against the number of demonstrations.}
    \vspace{-3mm}
    \label{fig:paraphase}
    \end{centering}
\end{wrapfigure}

Figure \ref{fig:paraphase} delineates the BLEU and Rouge-L scores for GPT and DTG in relation to various few-shot scenarios. In our preliminary experiments, we find that only 5-shot demonstrations cannot enable LLMs to clearly capture the underline mapping rule between the source and the target. To this end, we test LLMs on 20-shot and 50-shot and observe intriguing phenomenon.

Across all scenarios, DTG outperforms GPT models in terms of both BLEU and Rouge-L metrics. However, when the number of demonstrations is restricted, e.g., 1-shot and 5-shot, LLMs noticeably trail behind state-of-the-art systems. Interestingly, a significant enhancement in DTG performance is observed with the increase in the number of demonstrations. This improvement can be attributed to the model's enhanced ability to comprehend the underlying mapping rules between the source and target, a capability that intensifies with an expanded demonstration set.

\section{Analysis}
\label{sec:analysis}
In this section, we delve into a series of intriguing questions to elucidate the circumstances and reasons underpinning the robust performance of DTG. Unless specified otherwise, the base engine utilized throughout this investigation is \texttt{text-davinci-003}.

\paragraph{Ablation Study}

\begin{wraptable}[10]{r}{0.42\textwidth}
\vspace{-0.15in}
\caption{Ablations on DTG prompting.}
\label{tab:prompt_comparison}
\centering
\small
\setlength{\tabcolsep}{1.5pt}
\begin{tabular}{lcc}
\toprule
\bf Model & \bf BLEU &\bf COMET \\
\midrule
GPT 5-shot                  & 23.6 & 81.12\\
\; + DTG                    & 25.2 & 81.70\\
\; + w/o error detection    & 23.3 & 81.05\\ 
\; + wrong error type       & 25.3 & 81.74\\
\; + fixed error type       & 24.1 & 81.35\\
\; + correct candidate      & 23.0 & 81.17\\
\bottomrule
\end{tabular}
\label{tab:ablation_DTG}
\end{wraptable}

Prior research \cite{zhang2023prompting,vilar2022prompting,agrawal2022context} underscores the significant impact of both the quality and quantity of demonstrations on the performance of LLMs. Thus, it becomes essential to discern whether the improvements observed are attributable to modifications in the template or the deliberate capability inherent to the LLMs. To this end, we conduct experiments on WMT ZH-EN and show the comparisons in Table \ref{tab:ablation_DTG}. Firstly, eliminating the phrase ``Please detect the error type firstly, and refine the translation then'', denoted as ``w/o error detection'' in Table~\ref{tab:ablation_DTG}, DTG experiences a significant decrement in BLEU score, suggesting that the excised segment may contain crucial triggers stimulating the deliberate capability of the LLM. Along this line, we make two explorations: 1) replacing ``incorrect translation'' by ``good/correct translation'' in the demonstration only, resulting in no BLEU degradation, which is denoted as ``wrong error type'' in Table~\ref{tab:ablation_DTG}. This reveals that LLMs can rethink by themselves and make ``correct'' decisions though the demonstration is incorrect. 2) using fixed error type, e.g., under translation in the LLM response. This leads to a 1.1 BLEU drop, indicating that restricting the thought of LLMs would hinder the performance. Moreover, we observe that adopting the correct candidate generated by itself cannot bring further improvements than standard prompting. A plausible explanation for this observation could be that GPT3.5 might lack the necessary ability to accurately identify and concisely correct the parts of text requiring modification.

\begin{figure}
\centering
\begin{tikzpicture}

\begin{scope}[xshift=0em]
\begin{axis}[ 
width=3.9cm, height=3.5cm, 
ybar,
title={WMT De-En},
title style={anchor=center,yshift=0.8em},
xmin=0, xmax=10,
ymin=0, ymax=90,
ytick={0, 25, 50, 75},
major x tick style = transparent,
bar width=6pt,
xlabel={\scriptsize{GPT3.5\ \ \ \ GPT4}},
xlabel style={align=right,xshift=-0.0cm,yshift=0.4cm},
ylabel={Rate  (\%)},
ylabel style={yshift=-0.2cm,xshift=0cm},
enlarge x limits=0, 
x tick label style={opacity=0},
xtick=data,  
axis x line*=bottom,
axis y line*=left,
legend cell align=left,
anchor=south east,
column sep=1ex,
]  
\addplot[ybar, bar width=2.5mm, fill=bananayellow,bar shift=-1.6mm,  postaction={}] coordinates {(3, 29.8) (7, 38.4)};
\addplot[ybar, bar width=2.5mm, fill=blanchedalmond,bar shift=1.6mm,  postaction={}] coordinates {(3, 71.2) (7, 62.6)};  
\end{axis}
\end{scope}

\begin{scope}[xshift=1.2in]
\begin{axis}[
width=3.9cm, height=3.5cm,
ybar,
title={CNN/DailyMail},
title style={anchor=center,yshift=0.8em},
xmin=0, xmax=10,
ymin=0, ymax=90, 
ytick={0, 25, 50, 75},
major x tick style = transparent,
bar width=6pt,
xlabel={\scriptsize{GPT3.5\ \ \ \ GPT4}},
xlabel style={align=right,xshift=-0.0cm,yshift=0.4cm},
enlarge x limits=0, 
x tick label style={opacity=0},
xtick=data,  
axis x line*=bottom,
axis y line*=left,
legend cell align=left,
anchor=south east,
column sep=1ex,
]  
\addplot[ybar, bar width=2.5mm, fill=bananayellow,bar shift=-1.6mm,  postaction={}] coordinates {(3, 20) (7, 15)};
\addplot[ybar, bar width=2.5mm, fill=blanchedalmond,bar shift=1.6mm,  postaction={}] coordinates {(3, 80) (7, 85)};  
\end{axis}
\end{scope}

\begin{scope}[xshift=2.4in]
\begin{axis}[ 
width=3.9cm, height=3.5cm, 
ybar,
title={Asset},
title style={anchor=center,yshift=0.8em},
xmin=0, xmax=10,
ymin=0, ymax=90,
ytick={0, 25, 50, 75},
major x tick style = transparent,
bar width=6pt,
xlabel={\scriptsize{GPT3.5\ \ \ \ GPT4}},
xlabel style={align=right,xshift=-0.0cm,yshift=0.4cm},
enlarge x limits=0, 
x tick label style={opacity=0},
xtick=data,  
axis x line*=bottom,
axis y line*=left,
legend cell align=left,
anchor=south east,
column sep=1ex,
]  
\addplot[ybar, bar width=2.5mm, fill=bananayellow,bar shift=-1.6mm,  postaction={}] coordinates {(3, 11.2) (7, 23.9)};
\addplot[ybar, bar width=2.5mm, fill=blanchedalmond,bar shift=1.6mm,  postaction={}] coordinates {(3, 88.8) (7, 76.1)};    
\end{axis}
\end{scope}

\begin{scope}[xshift=3.6in]
\begin{axis}[ 
width=3.9cm, height=3.5cm, 
ybar,
title={EM},
title style={anchor=center,yshift=0.8em},
xmin=0, xmax=10,
ymin=0, ymax=90,
ytick={0, 25, 50, 75},
major x tick style = transparent,
bar width=6pt,
xlabel={\scriptsize{GPT3.5\ \ \ \ GPT4}},
xlabel style={align=right,xshift=-0.0cm,yshift=0.4cm},
enlarge x limits=0, 
x tick label style={opacity=0},
xtick=data,  
axis x line*=bottom,
axis y line*=left,
legend cell align=left,
anchor=south east,
column sep=1ex,
legend style={anchor=north west,font=\footnotesize,draw=none,column sep=1ex,xshift=-0.0em,yshift=0em,font=\footnotesize}
]  
\addplot[ybar, bar width=2.5mm, fill=bananayellow,bar shift=-1.6mm,  postaction={}] coordinates {(3, 48) (7, 52)};
\addplot[ybar, bar width=2.5mm, fill=blanchedalmond,bar shift=1.6mm,  postaction={}] coordinates {(3, 52) (7, 59)};  
\legend{Best, DTG}  
\end{axis}
\end{scope}

\end{tikzpicture}  
    \caption{GPT3.5 and GPT4 evaluation on 4 generation tasks. Note that we random select 500 samples due to the limitation of GPT4 access.}
    \label{fig:evaluation}
\end{figure}
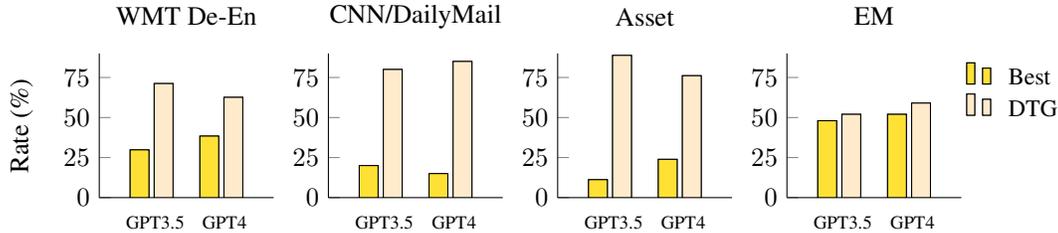

\paragraph{Evaluation by GPT Models}
\begin{table}
    \caption{Case study on refining from the previous candidate (Refine) and the proposed DTG method.}
    \label{tab:case_study}

    \small
    \centering
    \vspace{10pt}
    
    \begin{tabular}{lm{0.8\textwidth}}
        \toprule
        {Source} & \begin{CJK*}{UTF8}{gkai} 味道赞，肉类好，服务热情 \end{CJK*} \\
        {Reference} & {Nice taste, great meat, enthusiastic service.} \\
        
        {GPT 1-shot} &
        {The \colorbox{red!30}{taste is great}, the meat is \colorbox{red!30}{good}, and the service is \colorbox{red!30}{enthusiastic}.} \\

        {+ Refine} &
        {The \colorbox{red!30}{flavors are amazing}, the meat is \colorbox{red!30}{excellent}, and the service is \colorbox{red!30}{warm and welcoming}.} \\
        
        {+ DTG} &
        {\colorbox{green!30}{Great taste, good meat, enthusiastic service.}} \\

        \midrule
        {Source} & \begin{CJK*}{UTF8}{gkai}目前已经购买了这个系列3款机器！ \end{CJK*} \\
        {Reference} & {I have bought three laptops of this series!} \\
        
        {GPT 1-shot} &
        {\colorbox{red!30}{So far}, 3 machines from this series have been purchased!} \\

        {+ Refine} &
        {\colorbox{red!30}{Up until now}, 3 machines from this series have been purchased!} \\
        
        {+ DTG} &
        {\colorbox{green!30}{I have already purchased} 3 models from this series!} \\
        \bottomrule
    \end{tabular}
    \vspace{-10pt}
\end{table}

As previously discussed, despite DTG's impressive performance, it falls short of BART in some scenarios—most notably, it exhibits a significant gap in terms of Rouge scores in summarization tasks. However, \citet{liu2023gpteval} suggested that Rouge may not accurately represent the true performance of summarization tasks, given its poor alignment with human evaluations. In contrast, GPT models achieve optimal alignment with human justification and substantially outperform all previous state-of-the-art evaluators on the SummEval benchmark. This observation prompts an investigation into whether the generation output by DTG can surpass that of BART. Following their suggestion, we conduct reference-based evaluation and design a prompt as shown in Figure \ref{fig:evaluation}. We extract 500 test sets and compared DTG with the best result using GPT3.5 and GPT4 to select a better candidate. We see that DTG significantly beats the best system within GPT evaluation, except for the style transfer dataset.

\paragraph{Error Statistical Analysis}

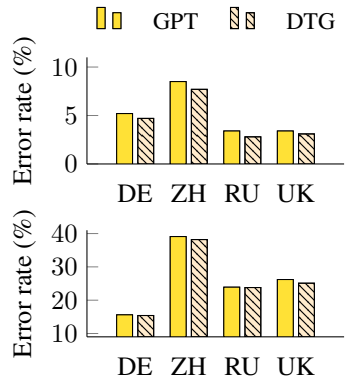
\begin{wrapfigure}[16]{r}{0.35\textwidth}
\vspace{-0.5cm}
\pgfplotsset{width=5.2cm, height=4.5cm}
    \centering
    \begin{tikzpicture}  
    \begin{scope}
        \begin{axis}  
        [  
            ybar,
            ymin=0, ymax=11,
            ytick={0, 5.0, 10.0},
            major x tick style = transparent,
            bar width=6pt,
            width=5cm,
            height=3cm,
            enlarge x limits=0.3,
            ylabel={Error rate (\%)},
            ylabel style={yshift=-0.4cm},
            symbolic x coords={DE, ZH, RU, UK},  
            xtick=data,  
            axis x line*=bottom,
            axis y line*=left,
            legend columns = 2,
            legend style={anchor=south,column sep=10pt,font=\small,draw=none,xshift=-1.6cm,yshift=0.2cm}
            ]  
        \addplot[ybar, fill=bananayellow,  postaction={}] coordinates {
            (DE, 5.2) (ZH, 8.5) (RU, 3.4) (UK, 3.4) 
        };
        \addplot[ybar, fill=blanchedalmond,  postaction={pattern=north west lines}] coordinates {
            (DE, 4.7) (ZH, 7.7) (RU, 2.8) (UK, 3.1) 
        };  
        \legend{GPT, DTG}  
        \end{axis}  
    \end{scope}

    \begin{scope}[yshift=-2.3cm]
        \begin{axis}  
        [  
            ybar,
            ymin=9, ymax=41,
            ytick={10.0, 20.0, 30.0, 40.0},
            major x tick style = transparent,
            bar width=6pt,
            width=5cm,
            height=3cm,
            enlarge x limits=0.3,
            ylabel={Error rate (\%)},
            ylabel style={yshift=-0.4cm},
            symbolic x coords={DE, ZH, RU, UK},  
            xtick=data,  
                axis x line*=bottom,
                axis y line*=left,
            ]  
        \addplot[ybar, fill=bananayellow,  postaction={}] coordinates {
            (DE, 15.6) (ZH, 39.1) (RU, 23.9) (UK, 26.2) 
        };
        \addplot[ybar, fill=blanchedalmond,  postaction={pattern=north west lines}] coordinates {
            (DE, 15.4) (ZH, 38.2) (RU, 23.8) (UK, 25.1) 
        };  
        \end{axis}  
    \end{scope}
    
    \end{tikzpicture}  
    \caption{
    Statistics of error rate for under translation (above) and entity translation (below).}
    \label{fig:error_analysis}
\end{wrapfigure}
To evaluate whether the proposed DTG prompting can facilitate error avoidance in GPT, we conduct error statistics on machine translation, where two frequently occurring error types are considered (i.e., under translation and incorrect entity translation)~\cite{hassan2018achieving}. Figure~\ref{fig:error_analysis} provides a comparison of the error rates between GPT models with and without the application of the DTG method. It is obvious to see that DTG reduces both error rates compared with the direct generation manner.

\paragraph{Case Study}
We provide a case study based on GPT4 model in Table~\ref{tab:case_study}, where ``Refine'' indicates utilizing the 5-shot baseline results as the synthesized sentences, i.e., ``[INCORRECT SYS]'' in Figure~\ref{fig:prompt_comparison}, and DTG is our method that uses an empty string instead. The conclusions are two-fold. 
1) Using the baseline results will cause the model to avoid generating the same segmentations in it although they may be correct already, e.g., ``taste'' to ``flavors'', ``so far'' to ``up until now'', as well as others in red. As a result, the fluency and accuracy of the final results may be affected.
2) Equipped with DTG, fluency, coherence and grammatical correctness of generated results are all promoted. In the first case, the DTG result is more faithful not only in semantics but also in structure than the baseline. In the second case, DTG is able to complete the subject ``I'' which does not appear in the source sentence.

\paragraph{DTG Can Serve as A Good Refiner}
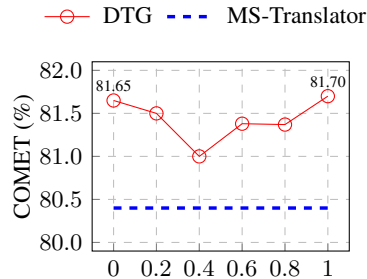
\begin{wrapfigure}[13]{r}{0.4\textwidth}
\pgfplotsset{}
\vspace{-0.35cm}
    \begin{centering}
    \begin{tikzpicture}
        \pgfplotsset{footnotesize,samples=2}
        \begin{groupplot}[
            group style = {group size = 1 by 1, horizontal sep = 40pt}, ]
            \nextgroupplot[
                align = center,
                legend style={at={(0.45,1.35)},anchor=north,draw=none,column sep=3pt},
                legend columns = 2,
                width=5.0cm, height=4.1cm,
                xmin=-0.1, xmax=1.1,
                ymin=79.9, ymax=82.1,
                xtick={0.0, 0.2, 0.4, 0.6, 0.8, 1.0},
                ylabel={COMET (\%)},
                ylabel style={align=center},
                ytick={80.0, 80.5, 81.0, 81.5, 82.0},
                xlabel style={at={(axis description cs:0.5,-0.15)},anchor=north},
                ylabel style={at={(axis description cs:0.17,0.5)},anchor=south},
                yticklabel style={/pgf/number format/precision=1,/pgf/number format/fixed zerofill},
                xtick pos=bottom,
                ytick pos=left,
                xmajorgrids,
                ymajorgrids,
                major grid style={dashed},
                ]
                \addplot[
                    color=red,
                    mark=o,
                    mark size=2.5pt,
                    ]
                    coordinates {
                    (0.0,  81.65)
                    (0.2,  81.50)
                    (0.4,  81.0)
                    (0.6,  81.38)
                    (0.8,  81.37)
                    (1.0,  81.7)
                    };
                    \addlegendentry{DTG}
                \addplot[
                    color=blue,
                    mark=square,
                    no markers,
                    dashed,
                    line width=1.3pt,
                    ]
                    coordinates {
                    (0,   80.4)
                    (1.0, 80.4)
                    };
                    \addlegendentry{MS-Translator}

                    \node[anchor=south,font=\tiny] at (axis cs:1., 81.7) {81.70};
                    \node[anchor=south,font=\tiny] at (axis cs:0., 81.65) {81.65};
            
        \end{groupplot}
    \end{tikzpicture}
    \caption{
    COMET \textit{v.s.} word drop rate of MS-Translator candidate.}
    \label{fig:drop_comet}
    \end{centering}
\end{wrapfigure}
To investigate the correlation between the performance of DTG and the provided candidate, we consider the translation task on WMT ZH-EN and create candidates with varying quality by randomly removing certain words from the translations generated by MS-Translator. Figure \ref{fig:drop_comet} displays the performance measured by COMET versus the word drop rate. The blue line represents the performance of MS-Translator, and the red line represents DTG with various candidates. Upon deliberation, GPT can improve the translation of MS-Translator from 80.4 to 81.65 in COMET. As aforementioned that DTG suffers from performance degradation when the candidate is a correct one generated by itself (See the last line in Table~\ref{tab:ablation_DTG}). However, upon deeper investigation, we discern that selecting candidates from systems other than GPT itself is a superior choice. This underscores the effectiveness of our DTG framework, demonstrating its capability to work even with high-quality candidates generated by other systems. Moreover, the performance declines when more words are dropped from the MS-Translator candidate, but interestingly, it increases when the candidate almost resembles an \emph{empty string}. Though with additional high-quality systems, DTG also successfully improves the performance, using an \emph{empty string} as a candidate can always lead to a better outcome without any additional resources and cost, as well as specific demonstration construction.

\section{Conclusions}

In this paper, we propose DTG prompting, which encourages LLMs to deliberate before generating the final results by letting the model detect the error type on a synthetic text that may contain errors. Using an empty string as the synthetic text successfully gets rid of an extra baseline system and improves the quality of the generated text. The DTG prompting can be easily applied to various text generation tasks with minimal adjustments in the prompt. Extensive experiments conducted on over 20 datasets across 7 text generation tasks demonstrate the effectiveness and broad applicability of the DTG prompting framework. One potential avenue for further enhancing the efficacy of DTG prompting involves leveraging task-specific domain knowledge. (e.g., explicitly listing the potential error types in the prompts to provide guidance for deliberation), which is worth future investigation.

\medskip

\bibliographystyle{plainnat}
\bibliography{neurips_2023}

\newpage
\appendix

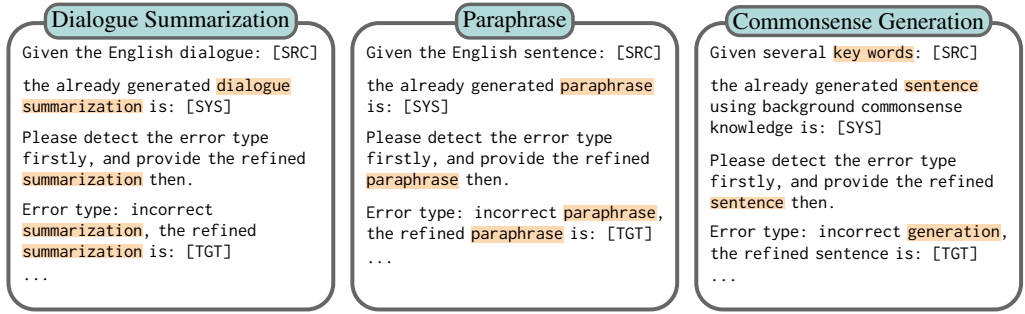
\begin{figure*}
    \centering

    \scriptsize{
\begin{tikzpicture}

\begin{scope}[]

\node [anchor=south west] (n1) at (0, 0) {};

\node [anchor=north,rectangle,rounded corners=10pt,minimum height=1.5in,minimum width=1.7in, draw, line width=1.5pt, draw=black!60!white] (b1) at ([xshift=0em, yshift=-1em]n1.south) {};

\node [anchor=center,rectangle,rounded corners=5pt,minimum height=1em,minimum width=2em,fill=teal!30, draw=black!60!white, line width=1.3pt] (l1) at ([xshift=0em, yshift=0em]b1.north) {\small{Dialogue Summarization}};

\node [anchor=north west,rectangle,rounded corners=2pt,minimum height=1em,minimum width=2em, text width=1.6in] (n21) at ([xshift=0.5em, yshift=-1em]b1.north west) {\texttt{Given the English dialogue: [SRC]}};
\node [anchor=north west,rectangle,rounded corners=2pt,minimum height=1em,minimum width=2em, text width=1.6in] (n22) at ([xshift=0em, yshift=-0.2em]n21.south west) {\texttt{the already generated \colorbox{orange!30}{dialogue} \colorbox{orange!30}{summarization} is: [SYS]}};
\node [anchor=north west,rectangle,rounded corners=2pt,minimum height=1em,minimum width=2em, text width=1.6in] (n23) at ([xshift=0em, yshift=-0.2em]n22.south west) {\texttt{Please detect the error type firstly, and provide the refined \colorbox{orange!30}{summarization} then.}};

\node [anchor=north west,rectangle,rounded corners=2pt,minimum height=1em,minimum width=2em, text width=1.6in] (n24) at ([xshift=0em, yshift=-0.2em]n23.south west) {\texttt{Error type: incorrect \colorbox{orange!30}{summarization}, the refined \colorbox{orange!30}{summarization} is: [TGT]}};
\node [anchor=north west,rectangle,rounded corners=2pt,minimum height=1em,minimum width=2em, text width=1.6in] (n25) at ([xshift=0em, yshift=-0.2em]n24.south west) {\texttt{...}};


\end{scope}

\begin{scope}[xshift=1.8in]

\tikzstyle{node1}=[rectangle,rounded corners=2pt,inner sep=0mm,minimum height=1.2em,minimum width=6em,fill=teal!45,font=\footnotesize]

\node [anchor=south west] (n1) at (0, 0) {};

\node [anchor=north,rectangle,rounded corners=10pt,minimum height=1.5in,minimum width=1.7in, draw, line width=1.5pt, draw=black!60!white] (b1) at ([xshift=0em, yshift=-1em]n1.south) {};

\node [anchor=center,rectangle,rounded corners=5pt,minimum height=1em,minimum width=2em,fill=teal!30, draw=black!60!white, line width=1.3pt] (l1) at ([xshift=0em, yshift=0em]b1.north) {\small{Paraphrase}};

\node [anchor=north west,rectangle,rounded corners=2pt,minimum height=1em,minimum width=2em, text width=1.6in] (n21) at ([xshift=0.5em, yshift=-1em]b1.north west) {\texttt{Given the English sentence: [SRC]}};
\node [anchor=north west,rectangle,rounded corners=2pt,minimum height=1em,minimum width=2em, text width=1.6in] (n22) at ([xshift=0em, yshift=-0.2em]n21.south west) {\texttt{the already generated \colorbox {orange!30}{paraphrase} is: [SYS]}};
\node [anchor=north west,rectangle,rounded corners=2pt,minimum height=1em,minimum width=2em, text width=1.6in] (n23) at ([xshift=0em, yshift=-0.2em]n22.south west) {\texttt{Please detect the error type firstly, and provide the refined \colorbox {orange!30}{paraphrase} then.}};

\node [anchor=north west,rectangle,rounded corners=2pt,minimum height=1em,minimum width=2em, text width=1.6in] (n24) at ([xshift=0em, yshift=-0.2em]n23.south west) {\texttt{Error type: incorrect \colorbox {orange!30}{paraphrase}, the refined \colorbox {orange!30}{paraphrase} is: [TGT]}};
\node [anchor=north west,rectangle,rounded corners=2pt,minimum height=1em,minimum width=2em, text width=1.6in] (n25) at ([xshift=0em, yshift=-0.2em]n24.south west) {\texttt{...}};
\end{scope}

\begin{scope}[xshift=3.6in]

\node [anchor=south west] (n1) at (0, 0) {};

\node [anchor=north,rectangle,rounded corners=10pt,minimum height=1.5in,minimum width=1.7in, draw, line width=1.5pt, draw=black!60!white] (b1) at ([xshift=0em, yshift=-1em]n1.south) {};

\node [anchor=center,rectangle,rounded corners=5pt,minimum height=1em,minimum width=2em,fill=teal!30, draw=black!60!white, line width=1.3pt] (l1) at ([xshift=0em, yshift=0em]b1.north) {\small{Commonsense Generation}};

\node [anchor=north west,rectangle,rounded corners=2pt,minimum height=1em,minimum width=2em, text width=1.6in] (n21) at ([xshift=0.5em, yshift=-1em]b1.north west) {\texttt{Given several \colorbox {orange!30}{key words}: [SRC]}};
\node [anchor=north west,rectangle,rounded corners=2pt,minimum height=1em,minimum width=2em, text width=1.6in] (n22) at ([xshift=0em, yshift=-0.2em]n21.south west) {\texttt{the already generated \colorbox {orange!30}{sentence} using background commonsense knowledge is: [SYS]}};
\node [anchor=north west,rectangle,rounded corners=2pt,minimum height=1em,minimum width=2em, text width=1.6in] (n23) at ([xshift=0em, yshift=-0.2em]n22.south west) {\texttt{Please detect the error type firstly, and provide the refined \colorbox {orange!30}{sentence} then.}};

\node [anchor=north west,rectangle,rounded corners=2pt,minimum height=1em,minimum width=2em, text width=1.6in] (n24) at ([xshift=0em, yshift=-0.2em]n23.south west) {\texttt{Error type: incorrect \colorbox {orange!30}{generation}, the refined sentence is: [TGT]}};
\node [anchor=north west,rectangle,rounded corners=2pt,minimum height=1em,minimum width=2em, text width=1.6in] (n25) at ([xshift=0em, yshift=-0.2em]n24.south west) {\texttt{...}};

\end{scope}

\end{tikzpicture}
}
\caption{Illustration of DTG demonstration design for dialogue summarization, paraphrase and commonsense generation tasks within minimal modifications.}
\label{fig:DTG_prompts}
\end{figure*}

\begin{figure*}
    \centering

    \scriptsize{
\begin{tikzpicture}

\begin{scope}[]

\node [anchor=south west] (n1) at (0, 0) {};

\node [anchor=north,rectangle,rounded corners=10pt,minimum height=1.5in,minimum width=1.7in, draw, line width=1.5pt, draw=black!60!white] (b1) at ([xshift=0em, yshift=-1em]n1.south) {};

\node [anchor=center,rectangle,rounded corners=5pt,minimum height=1em,minimum width=2em,fill=teal!30, draw=black!60!white, line width=1.3pt] (l1) at ([xshift=0em, yshift=0em]b1.north) {\small{Translation}};

\node [anchor=north west,rectangle,rounded corners=2pt,minimum height=1em,minimum width=2em, text width=1.8in] (n21) at ([xshift=0.5em, yshift=-1em]b1.north west) {\texttt{Given the [src] sentence: [SRC]}};
\node [anchor=north west,rectangle,rounded corners=2pt,minimum height=1em,minimum width=2em, text width=1.6in] (n22) at ([xshift=0em, yshift=-0.2em]n21.south west) {\texttt{the [tgt] \colorbox{orange!30}{translation} of the sentence is: [TGT]}};
\node [anchor=north west,rectangle,rounded corners=2pt,minimum height=1em,minimum width=2em, text width=1.6in] (n23) at ([xshift=0em, yshift=-0.2em]n22.south west) {\texttt{Given the [src] sentence: [Input]}};

\node [anchor=north west,rectangle,rounded corners=2pt,minimum height=1em,minimum width=2em, text width=1.6in] (n24) at ([xshift=0em, yshift=-0.2em]n23.south west) {\texttt{the [tgt] \colorbox{orange!30}{translation} of the sentence is:}};

\end{scope}

\begin{scope}[xshift=1.8in]

\tikzstyle{node1}=[rectangle,rounded corners=2pt,inner sep=0mm,minimum height=1.2em,minimum width=6em,fill=teal!45,font=\footnotesize]

\node [anchor=south west] (n1) at (0, 0) {};

\node [anchor=north,rectangle,rounded corners=10pt,minimum height=1.5in,minimum width=1.7in, draw, line width=1.5pt, draw=black!60!white] (b1) at ([xshift=0em, yshift=-1em]n1.south) {};

\node [anchor=center,rectangle,rounded corners=5pt,minimum height=1em,minimum width=2em,fill=teal!30, draw=black!60!white, line width=1.3pt] (l1) at ([xshift=0em, yshift=0em]b1.north) {\small{Dialogue Summarization}};

\node [anchor=north west,rectangle,rounded corners=2pt,minimum height=1em,minimum width=2em, text width=1.6in] (n21) at ([xshift=0.5em, yshift=-1em]b1.north west) {\texttt{Given the English \colorbox{orange!30}{dialogue}: [SRC]}};
\node [anchor=north west,rectangle,rounded corners=2pt,minimum height=1em,minimum width=2em, text width=1.6in] (n22) at ([xshift=0em, yshift=-0.2em]n21.south west) {\texttt{please summarize the main \colorbox{orange!30}{context}: [TGT]}};
\node [anchor=north west,rectangle,rounded corners=2pt,minimum height=1em,minimum width=2em, text width=1.6in] (n23) at ([xshift=0em, yshift=-0.2em]n22.south west) {\texttt{Given the English \colorbox{orange!30}{dialogue}: [Input]}};

\node [anchor=north west,rectangle,rounded corners=2pt,minimum height=1em,minimum width=2em, text width=1.6in] (n24) at ([xshift=0em, yshift=-0.2em]n23.south west) {\texttt{please summarize the main \colorbox{orange!30}{context}:}};
\end{scope}

\begin{scope}[xshift=3.6in]

\node [anchor=south west] (n1) at (0, 0) {};

\node [anchor=north,rectangle,rounded corners=10pt,minimum height=1.5in,minimum width=1.7in, draw, line width=1.5pt, draw=black!60!white] (b1) at ([xshift=0em, yshift=-1em]n1.south) {};

\node [anchor=center,rectangle,rounded corners=5pt,minimum height=1em,minimum width=2em,fill=teal!30, draw=black!60!white, line width=1.3pt] (l1) at ([xshift=0em, yshift=0em]b1.north) {\small{Simplification}};

\node [anchor=north west,rectangle,rounded corners=2pt,minimum height=1em,minimum width=2em, text width=1.6in] (n21) at ([xshift=0.5em, yshift=-1em]b1.north west) {\texttt{Given the English sentence: [SRC]}};
\node [anchor=north west,rectangle,rounded corners=2pt,minimum height=1em,minimum width=2em, text width=1.6in] (n22) at ([xshift=0em, yshift=-0.2em]n21.south west) {\texttt{the \colorbox{orange!30}{simplification} of the sentence is: [TGT]}};
\node [anchor=north west,rectangle,rounded corners=2pt,minimum height=1em,minimum width=2em, text width=1.6in] (n23) at ([xshift=0em, yshift=-0.2em]n22.south west) {\texttt{Given the English sentence: [Input]}};

\node [anchor=north west,rectangle,rounded corners=2pt,minimum height=1em,minimum width=2em, text width=1.6in] (n24) at ([xshift=0em, yshift=-0.2em]n23.south west) {\texttt{the \colorbox{orange!30}{simplification} of the sentence is:}};

\end{scope}

\begin{scope}[yshift=-1.7in]

\node [anchor=south west] (n1) at (0, 0) {};

\node [anchor=north,rectangle,rounded corners=10pt,minimum height=1.5in,minimum width=1.7in, draw, line width=1.5pt, draw=black!60!white] (b1) at ([xshift=0em, yshift=-1em]n1.south) {};

\node [anchor=center,rectangle,rounded corners=5pt,minimum height=1em,minimum width=2em,fill=teal!30, draw=black!60!white, line width=1.3pt] (l1) at ([xshift=0em, yshift=0em]b1.north) {\small{Style Transfer}};

\node [anchor=north west,rectangle,rounded corners=2pt,minimum height=1em,minimum width=2em, text width=1.6in] (n21) at ([xshift=0.5em, yshift=-1em]b1.north west) {\texttt{Given the English sentence: [SRC]}};
\node [anchor=north west,rectangle,rounded corners=2pt,minimum height=1em,minimum width=2em, text width=1.6in] (n22) at ([xshift=0em, yshift=-0.2em]n21.south west) {\texttt{please \colorbox{orange!30}{transfer} the style of the sentence into formal: [TGT]}};
\node [anchor=north west,rectangle,rounded corners=2pt,minimum height=1em,minimum width=2em, text width=1.6in] (n23) at ([xshift=0em, yshift=-0.2em]n22.south west) {\texttt{Given the English sentence: [Input]}};

\node [anchor=north west,rectangle,rounded corners=2pt,minimum height=1em,minimum width=2em, text width=1.6in] (n24) at ([xshift=0em, yshift=-0.2em]n23.south west) {\texttt{please \colorbox{orange!30}{transfer} the style of the sentence into formal:}};

\end{scope}

\begin{scope}[yshift=-1.7in, xshift=1.8in]

\node [anchor=south west] (n1) at (0, 0) {};

\node [anchor=north,rectangle,rounded corners=10pt,minimum height=1.5in,minimum width=1.7in, draw, line width=1.5pt, draw=black!60!white] (b1) at ([xshift=0em, yshift=-1em]n1.south) {};

\node [anchor=center,rectangle,rounded corners=5pt,minimum height=1em,minimum width=2em,fill=teal!30, draw=black!60!white, line width=1.3pt] (l1) at ([xshift=0em, yshift=0em]b1.north) {\small{Paraphrase}};

\node [anchor=north west,rectangle,rounded corners=2pt,minimum height=1em,minimum width=2em, text width=1.6in] (n21) at ([xshift=0.5em, yshift=-1em]b1.north west) {\texttt{Given the English sentence: [SRC]}};
\node [anchor=north west,rectangle,rounded corners=2pt,minimum height=1em,minimum width=2em, text width=1.6in] (n22) at ([xshift=0em, yshift=-0.2em]n21.south west) {\texttt{the \colorbox{orange!30}{paraphrase} of the sentence is: [TGT]}};
\node [anchor=north west,rectangle,rounded corners=2pt,minimum height=1em,minimum width=2em, text width=1.6in] (n23) at ([xshift=0em, yshift=-0.2em]n22.south west) {\texttt{Given the English sentence: [Input]}};

\node [anchor=north west,rectangle,rounded corners=2pt,minimum height=1em,minimum width=2em, text width=1.6in] (n24) at ([xshift=0em, yshift=-0.2em]n23.south west) {\texttt{the \colorbox{orange!30}{paraphrase} of the sentence is:}};

\end{scope}

\begin{scope}[yshift=-1.7in, xshift=3.6in]

\node [anchor=south west] (n1) at (0, 0) {};

\node [anchor=north,rectangle,rounded corners=10pt,minimum height=1.5in,minimum width=1.7in, draw, line width=1.5pt, draw=black!60!white] (b1) at ([xshift=0em, yshift=-1em]n1.south) {};

\node [anchor=center,rectangle,rounded corners=5pt,minimum height=1em,minimum width=2em,fill=teal!30, draw=black!60!white, line width=1.3pt] (l1) at ([xshift=0em, yshift=0em]b1.north) {\small{Commonsense Generation}};

\node [anchor=north west,rectangle,rounded corners=2pt,minimum height=1em,minimum width=2em, text width=1.6in] (n21) at ([xshift=0.5em, yshift=-1em]b1.north west) {\texttt{Given several \colorbox{orange!30}{key words}: [SRC]}};
\node [anchor=north west,rectangle,rounded corners=2pt,minimum height=1em,minimum width=2em, text width=1.6in] (n22) at ([xshift=0em, yshift=-0.2em]n21.south west) {\texttt{Please generate a coherent sentence using background \colorbox{orange!30}{commonsense knowledge} with the providing key words: [TGT]}};
\node [anchor=north west,rectangle,rounded corners=2pt,minimum height=1em,minimum width=2em, text width=1.6in] (n23) at ([xshift=0em, yshift=-0.2em]n22.south west) {\texttt{Given several \colorbox{orange!30}{key words}: [Input]}};

\node [anchor=north west,rectangle,rounded corners=2pt,minimum height=1em,minimum width=2em, text width=1.6in] (n24) at ([xshift=0em, yshift=-0.2em]n23.south west) {\texttt{Please generate a coherent sentence using background \colorbox{orange!30}{commonsense knowledge} with the providing key words:}};

\end{scope}
\end{tikzpicture}
}
\caption{Illustration of the standard GPT prompting involving both demonstration and test input on six generation tasks, including machine translation, dialogue summarization, text simplification, style transfer, paraphrase and commonsense generation.}
\label{fig:GPT_all}
\end{figure*}
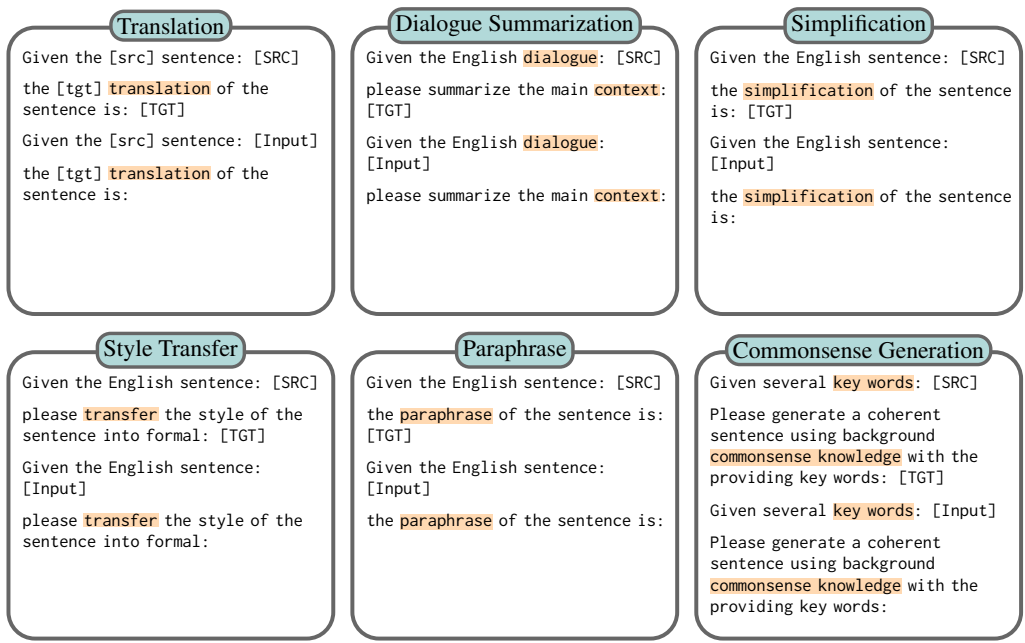

\section{Limitation}

Due to restricted access to GPT4, we have evaluated our \textit{Deliberate then Generate} (DTG) method on just two generation tasks: machine translation (across 8 language pairs) and simplification. There exists a necessity for more expansive experimentation across other tasks. Additionally, the effectiveness of DTG is contingent on model capacity. Models such as LLaMa-7B might not fully comprehend the instructions provided, resulting in weaker performance on downstream tasks. 
In our future work, we aim to ascertain the required scale of a language model to successfully facilitate deliberative generation.

Our work inherits the biases from pre-trained language models. For example, we only conduct experiments on English generation that GPT models are most powerful at. We provide results and analysis on English-to-Others translation in Appendix~\ref{sec:append_moreana}.
Future works could investigate the performance of DTG on multilingual pre-trained models.

\begin{figure*}
    \centering
    \footnotesize{
\begin{tikzpicture}

\begin{scope}[]

\node [anchor=south west] (n1) at (0, 0) {};

\node [anchor=north,rectangle,rounded corners=10pt,minimum height=1.3in,minimum width=5.0in, draw, line width=1.5pt, draw=black!60!white] (b1) at ([xshift=0em, yshift=-1em]n1.south) {};

\node [anchor=center,rectangle,rounded corners=5pt,minimum height=1em,minimum width=2em,fill=teal!30, draw=black!60!white, line width=1.3pt] (l1) at ([xshift=0em, yshift=0em]b1.north) {\small{Prompt template of GPT evaluation}};


\node [anchor=north west,rectangle,rounded corners=2pt,minimum height=1em,minimum width=2em, text width=4.8in] (n21) at ([xshift=0.5em, yshift=-1em]b1.north west) {\texttt{Given the [src] sentence: [SRC]}};
\node [anchor=north west,rectangle,rounded corners=2pt,minimum height=1em,minimum width=2em, text width=4.8in] (n22) at ([xshift=0em, yshift=-0.2em]n21.south west) {\texttt{Your task is to score the following two candidates translated by two systems, Candidate1: [sys1] Candidate2: [sys2].}};
\node [anchor=north west,rectangle,rounded corners=2pt,minimum height=1em,minimum width=2em, text width=4.8in] (n23) at ([xshift=0em, yshift=-0.2em]n22.south west) {\texttt{Please select the better one in terms of both coherence and fidelity. Note that C1 for Candidate1, C2 for Candidate2.}};

\node [anchor=north west,rectangle,rounded corners=2pt,minimum height=1em,minimum width=2em, text width=4.8in] (n24) at ([xshift=0em, yshift=-0.2em]n23.south west) {\texttt{Output:}};

\begin{pgfonlayer}{background}
\node [anchor=north,rectangle,rounded corners=3pt,minimum height=1.1in,minimum width=4.85in,fill=red!30] (bb) at ([xshift=0em, yshift=-0.2em]l1.south) {};
\node [anchor=north east,fill=red!30] (bl) at ([xshift=-0.2em, yshift=-0.2em]bb.north east) {\color{black}\textbf{Test}};
\end{pgfonlayer}
\end{scope}

\end{tikzpicture}
}
\caption{Illustration of the prompting design of GPT evaluation for Figure \ref{fig:evaluation}. We adhere to the recommendation proposed in \citep{liu2023gpteval}'s work, implementing a zero-shot GPT evaluation approach to identifying superior candidate translations through the adjudication of LLMs.}
\label{fig:prompt_evaluation}
\end{figure*}
\begin{figure*}
    \centering
    \footnotesize{
\begin{tikzpicture}

\begin{scope}[]

\node [anchor=south west] (n1) at (0, 0) {};

\node [anchor=north,rectangle,rounded corners=10pt,minimum height=1.4in,minimum width=5.0in, draw, line width=1.5pt, draw=black!60!white] (b1) at ([xshift=0em, yshift=-1em]n1.south) {};

\node [anchor=center,rectangle,rounded corners=5pt,minimum height=1em,minimum width=2em,fill=teal!30, draw=black!60!white, line width=1.3pt] (l1) at ([xshift=0em, yshift=0em]b1.north) {\small{(a) Prompt template of ``w/o error detection''}};

\node [anchor=north west,rectangle,rounded corners=2pt,minimum height=1em,minimum width=2em, text width=4.8in] (n21) at ([xshift=0.5em, yshift=-1em]b1.north west) {\texttt{Given the [src] sentence: [SRC]}};
\node [anchor=north west,rectangle,rounded corners=2pt,minimum height=1em,minimum width=2em, text width=4.8in] (n22) at ([xshift=0em, yshift=-0.2em]n21.south west) {\texttt{the [tgt] translation is: [SYS]}};
\node [anchor=north west,rectangle,rounded corners=2pt,minimum height=1em,minimum width=2em, text width=4.8in] (n23) at ([xshift=0em, yshift=-0.2em]n22.south west) {\texttt{The refined [tgt] translation is: [TGT]}};

\node [anchor=north west,rectangle,rounded corners=2pt,minimum height=1em,minimum width=2em, text width=4.8in] (n24) at ([xshift=0em, yshift=-0.2em]n23.south west) {\texttt{Given the [src] sentence: [SRC]}};
\node [anchor=north west,rectangle,rounded corners=2pt,minimum height=1em,minimum width=2em, text width=4.8in] (n25) at ([xshift=0em, yshift=-0.2em]n24.south west) {\texttt{the [tgt] translation is: [SYS]}};
\node [anchor=north west,rectangle,rounded corners=2pt,minimum height=1em,minimum width=2em, text width=4.8in] (n26) at ([xshift=0em, yshift=-0.2em]n25.south west) {\texttt{The refined [tgt] translation is:}};

\begin{pgfonlayer}{background}
\node [anchor=north,rectangle,rounded corners=3pt,minimum height=0.6in,minimum width=4.85in,fill=WindowsBlue!40] (bb) at ([xshift=0em, yshift=-0.2em]l1.south) {};
\node [anchor=north east,fill=WindowsBlue!40] (bl) at ([xshift=-0.2em, yshift=-0.2em]bb.north east) {\color{black}\textbf{Demonstration}};
\node [anchor=north,rectangle,rounded corners=3pt,minimum height=0.6in,minimum width=4.85in,fill=red!30] (br) at ([xshift=0em, yshift=-0.2em]bb.south) {};
\node [anchor=north east,fill=red!30] (bl) at ([xshift=-0.2em, yshift=-0.2em]br.north east) {\color{black}\textbf{Test}};
\end{pgfonlayer}

\end{scope}

\begin{scope}[yshift=-1.6in]

\node [anchor=south west] (n1) at (0, 0) {};

\node [anchor=north,rectangle,rounded corners=10pt,minimum height=1.8in,minimum width=5.0in, draw, line width=1.5pt, draw=black!60!white] (b1) at ([xshift=0em, yshift=-1em]n1.south) {};

\node [anchor=center,rectangle,rounded corners=5pt,minimum height=1em,minimum width=2em,fill=teal!30, draw=black!60!white, line width=1.3pt] (l1) at ([xshift=0em, yshift=0em]b1.north) {\small{(b) Prompt template of ``wrong error type''}};

\node [anchor=north west,rectangle,rounded corners=2pt,minimum height=1em,minimum width=2em, text width=4.8in] (n21) at ([xshift=0.5em, yshift=-1em]b1.north west) {\texttt{Given the [src] sentence: [SRC]}};
\node [anchor=north west,rectangle,rounded corners=2pt,minimum height=1em,minimum width=2em, text width=4.8in] (n22) at ([xshift=0em, yshift=-0.2em]n21.south west) {\texttt{the [tgt] translation is: [SYS]}};
\node [anchor=north west,rectangle,rounded corners=2pt,minimum height=1em,minimum width=2em, text width=4.8in] (n23) at ([xshift=0em, yshift=-0.2em]n22.south west) {\texttt{Please detect the error type firstly, and refine the translation then.}};
\node [anchor=north west,rectangle,rounded corners=2pt,minimum height=1em,minimum width=2em, text width=4.8in] (n24) at ([xshift=0em, yshift=-0.2em]n23.south west) {\texttt{Error type: \colorbox{orange!30}{good/correct translation}, the refined [tgt] translation is: [TGT]}};

\node [anchor=north west,rectangle,rounded corners=2pt,minimum height=1em,minimum width=2em, text width=4.8in] (n25) at ([xshift=0em, yshift=-0.2em]n24.south west) {\texttt{Given the [src] sentence: [SRC]}};
\node [anchor=north west,rectangle,rounded corners=2pt,minimum height=1em,minimum width=2em, text width=4.8in] (n26) at ([xshift=0em, yshift=-0.2em]n25.south west) {\texttt{the [tgt] translation is: [SYS]}};
\node [anchor=north west,rectangle,rounded corners=2pt,minimum height=1em,minimum width=2em, text width=4.8in] (n27) at ([xshift=0em, yshift=-0.2em]n26.south west) {\texttt{Please detect the error type firstly, and refine the translation then.}};
\node [anchor=north west,rectangle,rounded corners=2pt,minimum height=1em,minimum width=2em, text width=4.8in] (n28) at ([xshift=0em, yshift=-0.2em]n27.south west) {\texttt{Error type: }};

\begin{pgfonlayer}{background}
\node [anchor=north,rectangle,rounded corners=3pt,minimum height=0.8in,minimum width=4.85in,fill=WindowsBlue!40] (bb) at ([xshift=0em, yshift=-0.2em]l1.south) {};
\node [anchor=north east,fill=WindowsBlue!40] (bl) at ([xshift=-0.2em, yshift=-0.2em]bb.north east) {\color{black}\textbf{Demonstration}};
\node [anchor=north,rectangle,rounded corners=3pt,minimum height=0.8in,minimum width=4.85in,fill=red!30] (br) at ([xshift=0em, yshift=-0.2em]bb.south) {};
\node [anchor=north east,fill=red!30] (bl) at ([xshift=-0.2em, yshift=-0.2em]br.north east) {\color{black}\textbf{Test}};
\end{pgfonlayer}

\end{scope}

\begin{scope}[yshift=-3.6in]

\node [anchor=south west] (n1) at (0, 0) {};

\node [anchor=north,rectangle,rounded corners=10pt,minimum height=1.9in,minimum width=5.0in, draw, line width=1.5pt, draw=black!60!white] (b1) at ([xshift=0em, yshift=-1em]n1.south) {};

\node [anchor=center,rectangle,rounded corners=5pt,minimum height=1em,minimum width=2em,fill=teal!30, draw=black!60!white, line width=1.3pt] (l1) at ([xshift=0em, yshift=0em]b1.north) {\small{(c) Prompt template of ``fixed error type''}};

\node [anchor=north west,rectangle,rounded corners=2pt,minimum height=1em,minimum width=2em, text width=4.8in] (n21) at ([xshift=0.5em, yshift=-1em]b1.north west) {\texttt{Given the [src] sentence: [SRC]}};
\node [anchor=north west,rectangle,rounded corners=2pt,minimum height=1em,minimum width=2em, text width=4.8in] (n22) at ([xshift=0em, yshift=-0.2em]n21.south west) {\texttt{the [tgt] translation is: [SYS]}};
\node [anchor=north west,rectangle,rounded corners=2pt,minimum height=1em,minimum width=2em, text width=4.8in] (n23) at ([xshift=0em, yshift=-0.2em]n22.south west) {\texttt{Please detect the error type firstly, and refine the translation then.}};
\node [anchor=north west,rectangle,rounded corners=2pt,minimum height=1em,minimum width=2em, text width=4.8in] (n24) at ([xshift=0em, yshift=-0.2em]n23.south west) {\texttt{Error type: \colorbox{orange!30}{under translation}, the refined [tgt] translation is: [TGT]}};

\node [anchor=north west,rectangle,rounded corners=2pt,minimum height=1em,minimum width=2em, text width=4.8in] (n25) at ([xshift=0em, yshift=-0.2em]n24.south west) {\texttt{Given the [src] sentence: [SRC]}};
\node [anchor=north west,rectangle,rounded corners=2pt,minimum height=1em,minimum width=2em, text width=4.8in] (n26) at ([xshift=0em, yshift=-0.2em]n25.south west) {\texttt{the [tgt] translation is: [SYS]}};
\node [anchor=north west,rectangle,rounded corners=2pt,minimum height=1em,minimum width=2em, text width=4.8in] (n27) at ([xshift=0em, yshift=-0.2em]n26.south west) {\texttt{Please detect the error type firstly, and refine the translation then.}};
\node [anchor=north west,rectangle,rounded corners=2pt,minimum height=1em,minimum width=2em, text width=4.8in] (n28) at ([xshift=0em, yshift=-0.2em]n27.south west) {\texttt{Error type: \colorbox{orange!30}{under translation}, the refined [tgt] translation is:}};

\begin{pgfonlayer}{background}
\node [anchor=north,rectangle,rounded corners=3pt,minimum height=0.8in,minimum width=4.85in,fill=WindowsBlue!40] (bb) at ([xshift=0em, yshift=-0.2em]l1.south) {};
\node [anchor=north east,fill=WindowsBlue!40] (bl) at ([xshift=-0.2em, yshift=-0.2em]bb.north east) {\color{black}\textbf{Demonstration}};
\node [anchor=north,rectangle,rounded corners=3pt,minimum height=0.85in,minimum width=4.85in,fill=red!30] (br) at ([xshift=0em, yshift=-0.2em]bb.south) {};
\node [anchor=north east,fill=red!30] (bl) at ([xshift=-0.2em, yshift=-0.2em]br.north east) {\color{black}\textbf{Test}};
\end{pgfonlayer}
\end{scope}

\end{tikzpicture}
}
\caption{Illustration of the prompting design of the ablation study in Table \ref{tab:ablation_DTG}. Note that all [SYS] here is \emph{empty string}. The purpose here is to evaluate the deliberation ability of LLMs.}
\label{fig:prompt_ablations}
\end{figure*}

\section{Design of Prompts}
Figure \ref{fig:DTG_prompts} presents the DTG demonstration design across the other three text generation tasks. It can be observed that DTG does not necessitate task-specific designs; instead, a clear instruction outlining the main task for each work suffices. For the ease of replication of our results, we also furnish all baseline prompts, as depicted in Figure \ref{fig:GPT_all}. Also, we provide the prompting design for GPT evaluation in Figure \ref{fig:prompt_evaluation}, which follows a zero-shot fashion.

To facilitate a more comprehensive understanding of the prompt ablations conducted in Section \ref{sec:analysis}, we provide the corresponding design of prompts in Figure \ref{fig:prompt_ablations}. Please note that prompts in blue represent the pre-designed demonstration, while those in red represent the test input.
As observed, firstly, removing the error detection leads to the prompting in \ref{fig:prompt_ablations} (a). Additionally, the term ``wrong error type'' implies that we fed an \emph{empty string} into LLMs, presenting it as a good translation. However, LLMs can autonomously detect the correct error type as an ``incorrect translation'' and subsequently generate an accurate response following careful deliberation (Figure \ref{fig:prompt_ablations} (b)). Conversely, if we constrain the error type detection process and solely allow LLMs to generate the translation, a considerable performance gap emerges (See Figure \ref{fig:prompt_ablations} (c)).

\begin{table}[!t]
\centering
\caption{Evaluation results of GPT on six high-resource and two-low resource machine translation tasks from WMT Testsets in from English directions. The best scores are marked in bold.}
\label{tab:en2xx}
\resizebox{\textwidth}{!}{
\begin{tabular}{l|c c c c | c c c c}
\hline
\bf{System} &  \bf{COMET-22$\uparrow$}  & \bf{TER$\downarrow$} & \bf{ChrF$\uparrow$}  & \bf{BLEU$\uparrow$} &
\bf{COMET-22$\uparrow$}  & \bf{TER$\downarrow$} & \bf{ChrF$\uparrow$}  & \bf{BLEU$\uparrow$} \\ 
\hline
& \multicolumn{4}{c|}{EN-DE} &   \multicolumn{4}{c}{EN-ZH} \\ 

WMT-Best\dag            & \textbf{87.2} & \bf 49.9 &	\textbf{64.6} &	\textbf{38.4}       & \bf 86.7	& 102.3	 & 41.1 &	44.8  \\
MS-Translator\dag       & 86.8     & 50.5   & 64.2     & 37.3      & 86.1    & \bf 94.2	& \textbf{43.1} & 	\textbf{48.1}      \\
GPT 5-shot              &	86.3 & 54.6 &	61.3 & 33.3      &  \bf 86.7	 & 97.4 &	40.0 & 43.7     \\
\rowcolor{gray!40} 
\; + DTG                &	86.3 & 54.1 &	61.6 & 33.4      &	86.6    & 98.6 &	39.4 & 43.5      \\
\hline

& \multicolumn{4}{c|}{EN-CS} &   \multicolumn{4}{c}{EN-RU} \\ 
WMT-Best\dag            & \textbf{91.9} &	\textbf{43.7} &	\textbf{68.2} &	\textbf{45.8} &  \textbf{89.5} &	56.8 &	\textbf{58.3}	& 32.4 \\
MS-Translator\dag       & 90.6	& 45.7	& 65.6	& 42.1  & 87.4&	\bf 56.7 &	58.1&	\textbf{33.1}  \\
GPT 5-shot              &	88.9 & 54.6 &	58.9 & 32.7 &87.0 & 61.3 &	54.4 & 28.2 \\
\rowcolor{gray!40} 
\; + DTG                &	88.8 & 54.5 &	59.0 & 32.9  &	85.7 & 63.0 &	52.1 & 28.1 \\
\hline

& \multicolumn{4}{c|}{EN-JA} &   \multicolumn{4}{c}{EN-UK} \\ 
WMT-Best\dag            &\bf{89.3}	 & 	\bf 105.9	 & 	\textbf{36.8} &		\textbf{27.6}  & \textbf{88.8}	& \bf  57.5 &	\textbf{59.3}	& \textbf{32.5} \\
MS-Translator\dag       & 88.0	& 106.0 &	34.9 & 25.1  &  86.1 & 63.2	&   56.1	& 28.2 \\ 
GPT 5-shot              & 88.1  & 111.8 &	31.0 & 21.4  &	85.4 & 70.2 &   50.6    & 21.8 \\
\rowcolor{gray!40} 
\; + DTG                & 88.0  & 111.8 &	31.0 & 21.7  &	83.8 & 71.6 &	47.8    & 20.8 \\
\hline

& \multicolumn{4}{c|}{EN-IS} &   \multicolumn{4}{c}{EN-HA} \\
WMT-Best\dag            & \textbf{86.8} &	\bf 55.0	& \textbf{59.6} &	\textbf{33.3} & \textbf{79.8}&	\bf 65.6 &	\textbf{51.1} &	\textbf{20.1} \\
MS-Translator\dag       & 84.3   & 57.2 &	56.8 &	28.7  & 72.5  & 75.6 &	38.4 & 10.3\\

GPT 5-shot              &	76.1 & 70.8 &	44.1 & 16.2   &	72.8  & 87.4 &	38.5 & 9.9 \\
\rowcolor{gray!40} 
\; + DTG                &	76.7 & 70.9 &	44.2 & 16.3   &	73.2  & 77.7 &	39.3 & 10.1\\
\hline

\end{tabular}}
\end{table}

\section{More Analyses}
\label{sec:append_moreana}

\paragraph{Results on Machine Translation from English}
Table \ref{tab:en2xx} summarizes the results of standard prompting and our DTG method in 5-shot scenarios, alongside results from WMT-Best and MS-Translator. When compared to results from to-English directional language pairs, such as DE-EN, the improvements provided by DTG over the standard prompting strategy appear somewhat marginal. Furthermore, DTG may yield results inferior to standard prompting in EN-ZH and EN-UK scenarios. This can likely be ascribed to the disparities in the balance of training sets across different languages.

\paragraph{Ablations on Candidates}

\begin{wraptable}[9]{r}{0.45\textwidth}
\vspace{-0.25in}
\caption{Ablations on DTG prompting in terms of different candidates.}
\label{tab:prompt_comparison}
\centering
\small
\setlength{\tabcolsep}{1.5pt}
\begin{tabular}{clcc}
\toprule
\bf \#    &\bf Model & \bf BLEU &\bf COMET \\
\midrule
1   &  GPT 5-shot                        & 23.6 & 81.12\\
2   &\; + DTG                            & 25.2 & 81.70\\
3   &\; + fixed incorrect candidate      & 25.0 & 81.72\\ 
4   &\; + irrelevant languages           & 25.1 & 81.81\\
5   &\; + correct candidate      & 23.0 & 81.17\\
\bottomrule
\end{tabular}
\label{tab:ablation_prompt}
\end{wraptable}
In Section \ref{sec:method}, we demonstrated that the \emph{empty string} serves as a universal and effective choice to stimulate LLMs to engage in the \textit{Deliberate then Generate} process, and that a candidate more distinct from the reference can yield superior results. In this section, we aim to explore if other candidates may also prove effective in this context.
Here, we take the WMT ZH-EN  translation as an instance. Table \ref{tab:ablation_prompt} shows the comparison of various candidate inputs. Specifically, the term "fixed incorrect candidate" (\#3) refers to the use of a fixed yet incorrect (irrelevant) English translation as the candidate.\footnote{We random sample an English sentence: [SYS]: \textit{EBA Education Team together with Accace Ukraine invite you to join the EBA Education Update: Performance Audit.}} Likewise, system \#4 indicates that the candidates neither belong to the target language nor conform to the correct structure or grammar.\footnote{Similarly, we random sample an Ukraine sentence: [SYS]: \foreignlanguage{russian}{З впевненістю можете довіряти нам і будь ласка, звертайтеся до нас, якщо у вас є які-небудь питання чи коментарі.}} Interestingly, both 2 systems deliver comparable performance with our default setting, with system \#4 even achieving a higher COMET score. However, when shifting to a correct candidate, LLMs seem to underperform. This observation suggests that LLMs can effectively deliberate when the candidate is incorrect - whether it is an \emph{empty string} or other incorrect translations - and subsequently generate a substantially improved translation.

\section{Details of Datasets}
In this section, we offer more detailed statistics concerning the test sets utilized in this study, encompassing 8 machine translation, 4 summarization, 4 style transfer, 2 simplification, 1 commonsense generation, and 1 paraphrase benchmarks. Table \ref{tab:dataset} provides a summary of the number of test sets, total words, and the average length. We will release the test sets and the corresponding demonstrations in the future. Note that the statistic is conducted based on tokenization sequences, which would be further segmented by BPE before feeding into LLMs. Consequently, the average length of summarization inputs would appear significantly larger, leading to an elevated risk in the context of few-shot requests.

 \begin{table*}[t]
  \setlength{\tabcolsep}{2.0pt}
  \small
  \caption{Statistics of the dataset we used on over 20 benchmarks.  Note that ``Num.'' represents the number of test sets for each benchmark. ``Total Words'' and ``Ave. Words'' denote the total word count and average lengths, respectively. These statistics are based on tokenization sequences.}
  \centering
  \begin{tabular}{lccc|lccc}
  \midrule
  \bf Dataset & \bf Num. & \bf {Total Words} & \bf {Ave. Words}  & \bf Dataset & \bf Num. & \bf {Total Words} & \bf {Ave. Words} \\
  \midrule
  WMT DE-EN       & 1984 &    33540  & 16.9   &  CNN/DailyMail   & 11490    & 9017116        &   784.8                        \\
  WMT CS-EN       & 1448 &    26050  & 17.9   &  GigaWord        & 1951     & 72171          &   37.0                         \\
  WMT JA-EN       & 2008 &    36731      &  18.3     &  SamSum          & 819      & 104492         &   127.6                        \\
  WMT ZH-EN       & 1875 &    14353  & 7.7    &  DialogSum       & 500      & 96385          &   192.7                        \\
  WMT RU-EN       & 2016 &    32992  & 16.3   &  EM              & 1416     & 17279          &   12.2                         \\
  WMT UK-EN       & 2018 &    29273  & 14.5   &  FR              & 1332     & 16799          &   12.6                         \\
  WMT IS-EN       & 1000 &    19930      &  19.9     &  Amazon          & 500      & 6055           &   12.1                         \\
  WMT HA-EN       & 997  &    30955  & 31.0   &  Yelp            & 500      & 5432           &   10.9                         \\
  CommonGen       & 993  &    6465   & 6.5    &  Asset           & 359      & 8115           &   22.6                         \\
  QQP             & 2500 &    27543  & 11.0   &  Wiki-auto       & 2000     & 43860          &   21.9                         \\
  \bottomrule

\end{tabular}
  \label{tab:dataset}
\end{table*}

\section{Details of Error Statistical}
In Figure~\ref{fig:error_analysis}, two types of error are considered~(i.e., under translation and entity translation error). In this section, we provide the details of the method to conduct the error statistics.

\paragraph{Under Translation} We first use \emph{awesome-align}\footnote{https://github.com/neulab/awesome-align} to get the alignment between the source and target sentences. Then, a word in the source sentence is regarded as under translation, when it is aligned to a word in the reference target sentence but failed to be aligned in the generated target sentence.

\paragraph{Entity Translation} We first use \emph{spaCy}\footnote{https://github.com/explosion/spaCy} to recognize the named entities in the reference target sentence, where person names, organizations and locations are considered. Then, an entity in the reference is considered an error if it cannot be found in the generated target sentence.

\end{document}